\documentclass{article}

\PassOptionsToPackage{numbers,sort&compress,comma}{natbib}
\usepackage{arxiv}

\usepackage[utf8]{inputenc}
\usepackage[T1]{fontenc}
\usepackage{hyperref}
\usepackage{url}
\usepackage{booktabs}
\usepackage{amsfonts}
\usepackage{amssymb}
\usepackage{nicefrac}
\usepackage{microtype}
\usepackage{enumitem}
\usepackage{graphicx}
\usepackage{doi}
\usepackage{amsmath}
\usepackage{xcolor}
\usepackage{subcaption}
\usepackage{url}
\usepackage{multirow}
\usepackage{amsmath}
\usepackage{multirow}
\usepackage{graphicx}
\usepackage{subcaption}
\usepackage{mathtools}
\usepackage{amsthm}
\usepackage{enumitem}
\usepackage{microtype}
\usepackage{wrapfig}
\usepackage{rotating}
\usepackage{makecell}
\usepackage{algorithm}
\usepackage{algpseudocode}

\title{
BoRAD: Bootstrap your Own Representations for Multi-class Anomaly Detection
}

\author{
\href{mailto:duyhkse184883@fpt.edu.vn}{Hoang Khuong Duy} \\
Department of Artificial Intelligence\\
FPT University\\
Ho Chi Minh City, Vietnam\\
\texttt{luythoangduy@gmail.com}
\And
\href{mailto:minhtringuyen130205@gmail.com}{Nguyen Minh Tri} \\
Department of IT\\
FPT University\\
Ho Chi Minh City, Vietnam\\
\texttt{minhtringuyen130205@gmail.com}
\And
\href{mailto:nguhcv@fe.edu.vn}{Huynh Cong Viet Ngu} \\
Department of Computing Fundamental\\
FPT University\\
Ho Chi Minh City, Vietnam\\
\texttt{nguhcv@fe.edu.vn}
}

\renewcommand{\shorttitle}{\textit{arXiv} Template}

\hypersetup{
pdftitle={A template for the arxiv style},
pdfsubject={q-bio.NC, q-bio.QM},
pdfauthor={David S.~Hippocampus, Elias D.~Striatum},
pdfkeywords={First keyword, Second keyword, More},
}

\begin{document}

\maketitle

\begin{abstract}
    Reconstruction-based anomaly detection is attractive for industrial inspection, but scaling it from category-specific training to a one-for-all setting is challenging. A single model must reconstruct diverse normal appearances without copying abnormal details, which exposes two coupled failure modes: identical shortcut, where anomalies pass through the reconstruction path, and mis-reconstruction, where normal categories are confused with one another. We propose \textbf{BoRAD}, a label-free training framework that treats this as a representation-capacity allocation problem. BoRAD uses a shared learnable prototype bank to impose two complementary regularizers: spatial prototype alignment contracts local within-prototype variation to suppress anomaly copying, while prototype-relative global alignment preserves between-prototype structure and improves sensitivity to abnormal angular deviations. The prototype bank and prediction heads are used only during training; inference remains a standard teacher-student feature discrepancy pass, with no class labels, negative pairs, memory retrieval, or prototype lookup. BoRAD achieves competitive one-for-all anomaly detection performance, including 86.2\% mAD on MVTec AD, 80.7\% mAD on VisA and 73.1\% mAD on Real-IAD. Diagnostic analyses further show reduced anomaly leakage, improved normal-category separability, and stronger anomaly-normal score separation.
\end{abstract}

\section{Introduction}\label{sec:intro}

Unsupervised anomaly detection plays an important role in high-reliability domains such as medical diagnosis~\cite{han2021madgan, gonzalez2025hyperbolic} and industrial inspection~\cite{liu2024deep,roth2022towards}, where abnormal patterns are rare but can indicate critical diseases or manufacturing defects. Reconstruction-based industrial anomaly detection (IAD) learns normal visual patterns and identifies anomalies through reconstruction discrepancy at test time~\cite{liu2024deep,deng2022anomaly,zong2018deep,schlegl2019f,gudovskiy2022cflow}. In the conventional one-for-one setting, a separate detector is trained for each object category~\cite{roth2022towards,gong2019memorizing,chen2024unified}. This specialization often works well, but it is costly to scale: practical deployments may contain many categories, changing product lines, and limited tolerance for maintaining category-specific models or inference routing~\cite{you2022unified}.

A more scalable alternative is multi-class, or one-for-all, anomaly detection, where a single model is trained on normal samples pooled from multiple categories~\cite{you2022unified,fan2025salvaging}. This setting changes the nature of reconstruction because the model no longer needs to represent a single compact normal distribution, but must reconstruct the normal appearances of many categories at once. This induces two distinct failure modes. First, \emph{mis-reconstruction} arises when category-specific normal structures are not sufficiently separated in the representation: the decoder may reconstruct an input from one category using the appearance of another, producing inter-class confusion~\cite{fan2025salvaging}. Second, \emph{identical shortcut} occurs when the reconstruction path learns a near-identity mapping and preserves input details too faithfully, allowing abnormal regions to be reconstructed as normal~\cite{you2022adtr,you2022unified}. This shortcut can appear in reconstruction models in general, but the multi-class setting makes it more severe because the reconstructor must cover a much broader normal distribution, making input copying an attractive optimization strategy.

Prior work has addressed these two problems from complementary directions. To reduce mis-reconstruction, class-conditioned modules, prompts, memory mechanisms, and class-aware objectives improve category-level discriminability~\cite{yao2023one,lv2025one,li2024promptad,park2020learning,jeong2023winclip,lu2023hierarchical,fan2025salvaging}. To reduce identical shortcut, masking-based methods hide partial visual evidence to prevent direct copying~\cite{lu2026maskad}, attention-based reconstructors avoid overly direct local mappings~\cite{you2022adtr}, and memory or quantization mechanisms compress the latent representation so that it differs more strongly from the raw input~\cite{park2020learning,lu2023hierarchical}. However, small masking ratios impose only weak constraints, while many discriminability-oriented methods require class supervision, pseudo labels, negative pairs, or inference-time memory.

These observations suggest that both failure modes are representation problems: the latent space must suppress input-copying factors while preserving category-specific normal structure. This is consistent with prior latent-space interventions in anomaly detection and with representation-learning studies linking shortcut behavior to which factors are preserved or suppressed~\cite{oord2018representation,wang2021dense,robinson2021can}. Based on this view, we propose \textbf{BoRAD} (Bootstrap your Own Representations for Anomaly Detection), a label-free framework that uses a shared learnable prototype bank to shape the reconstruction representation during training. A spatial prototype alignment path contracts local within-prototype variation to reduce identical shortcuts, while a global coordinate-shift path preserves between-prototype structure to avoid mis-reconstruction. The prototype bank and prediction heads are used only during training; inference remains a standard teacher-student reconstruction pass without prototype lookup, clustering, or memory retrieval.

To examine whether these mechanisms address the above failure modes, Section~\ref{sec:diagnostic_analysis} reports diagnostics for anomaly leakage, normal-category separability, and anomaly-normal score separation beyond aggregate benchmark scores.

Our contributions are summarized as follows:
\begin{itemize}[nosep]
\item \textbf{A representation-dilemma view of one-for-all AD.} We formulate multi-class reconstruction through two coupled failure modes, identical shortcut and mis-reconstruction, and relate them to bottleneck and variance decomposition principles~\cite{kawaguchi2023does}.
\item \textbf{Prototype-guided representation bifurcation.} We introduce a shared prototype bank incorporate with contrastive that contracts local within-prototype variation while preserving global discriminability, without class labels, negative pairs, or inference-time memory.
\item \textbf{Training-only regularization with strong standard inference.} BoRAD uses prototype alignment only during training; at test time, anomaly maps are computed by teacher-student feature discrepancy and achieve competitive one-for-all IAD performance.
\end{itemize}

% The remainder of the paper is organized as follows. Section~\ref{sec:related_work} reviews prior work on unsupervised anomaly detection, unified anomaly detection under reconstruction shortcuts, and representation learning perspectives on shortcut behavior. Section~\ref{sec:method} then motivates BoRAD from the representation dilemma in unified anomaly detection, explains how shortcut-dominated reconstruction features arise, and introduces the proposed training objective based on dense prototype alignment and shifted global residuals. Section~\ref{sec:experimental_results} reports benchmark comparisons, ablation studies, qualitative results, and diagnostic analyses. The appendices provide implementation details, theoretical interpretations, additional ablations, and full benchmark tables.

\section{Related Work}\label{sec:related_work}

\textbf{Unsupervised Anomaly detection methods.}
Anomaly detection aims to identify samples or regions that deviate from the normal data distribution. Existing methods~\cite{liu2024deep} can be broadly divided into three categories: (1) Reconstruction-based methods that learn to reconstruct normal samples and detect anomalies through reconstruction failures~\cite{you2022unified, bergmann2018improving, yang2020dfr, he2024diffusion, he2024mambaad, zhang2025exploring}; (2) Distillation-based methods that train a student network to reconstruct the representations of a frozen teacher network using normal samples, and localize anomalies by teacher--student feature discrepancies~\cite{deng2022anomaly, zhang2023destseg, feng2025omiad, liu2025unlocking}; (3) Embedding-based methods that model normal feature distributions in a representation space and detect anomalies according to feature distance, density, or deviation from normal clusters~\cite{roth2022towards, zavrtanik2021draem, liu2023simplenet}. Most existing methods rely on ImageNet-pretrained~\cite{deng2009imagenet} networks for feature extraction, as these models provide rich and transferable representations. However, industrial images often exhibit an inevitable distribution shift from natural images, which can limit the suitability of directly using pretrained features. To address this issue, recent methods introduce additional modules for feature adaptation~\cite{gu2024rethinking}, representation learning~\cite{reiss2023mean, guo2023recontrast}, or compact feature transformation~\cite{lu2023hierarchical}. Memory-based mechanisms~\cite{gong2019memorizing} are also commonly used to store and retrieve representative normal patterns, enabling the model to better capture the normal data distribution.

\textbf{Unified anomaly detection and Key challenges.}
Unified anomaly detection faces two key challenges: \emph{mis-reconstruction} and \emph{identical shortcut}.

\emph{Mis-reconstruction.}
Mis-reconstruction arises when a model trained on mixed normal data reconstructs an input from one category using patterns from another, leading to inter-class confusion~\cite{fan2025salvaging}. Prior methods mitigate this issue with class-conditioned modules~\cite{yao2023one}, prompts~\cite{lv2025one,li2024promptad}, memory mechanisms~\cite{park2020learning}, or class-aware objectives~\cite{jeong2023winclip,lu2023hierarchical,fan2025salvaging}. These methods generally make representations more diverse and discriminative at the category level.

\emph{Identical shortcut.}
Identical shortcut occurs when the model learns a near-identity mapping and faithfully reconstructs both normal and anomalous regions, thereby weakening anomaly signals. ADTR~\cite{you2022adtr} attributes this behavior to the ease with which CNNs can learn identity mappings, e.g., a $1\times1$ convolution becomes $x'=x$ when $W=I$ and $b=0$, and argues that attention-based reconstructors may alleviate it by relying on token interactions instead of direct local mappings. In unified anomaly detection, multi-class reconstruction further increases the difficulty of learning category-specific normal patterns, making identical shortcut an attractive solution~\cite{you2022unified}. Masking-based methods therefore hide partial visual information and force the model to infer normal structures instead of simply copying the input, with methods such as~\cite{lu2026maskad, yao2023one, feng2025omiad} further improving the masking strategy. A complementary line of work views identical shortcut as a latent-space problem: memory-based methods constrain latent features through normal memory retrieval~\cite{park2020learning}, while quantization-based approaches compress the latent space so that only normal patterns are preserved~\cite{lu2023hierarchical}.

\textbf{Representation learning and shortcut behavior.}
Representation learning aims to encode the factors of variation that are useful for downstream tasks~\cite{bengio2013representation}. Shortcut learning can be viewed as a representation failure: the model preserves easily exploitable or spuriously correlated factors while suppressing more robust factors~\cite{saranrittichai2022overcoming,robinson2021can}. This view is useful for one-for-all anomaly detection, where the latent space must both preserve category-specific normal factors and prevent input-copying factors from dominating reconstruction. Contrastive learning provides one example of this trade-off: class-aware contrastive objectives can improve category separability~\cite{fan2025salvaging}, but instance discrimination can also rely on shortcut features and suppress other useful factors~\cite{robinson2021can}. BoRAD therefore adopts a negative-free BYOL-style objective~\cite{grill2020bootstrap}, but applies it through dense prototype alignment and shifted global residuals to shape reconstruction features without class labels or negative pairs.

\section{Theoretical Framework}
\label{sec:method}

\begin{figure}[h]
    \centering
    \includegraphics[width=\textwidth]{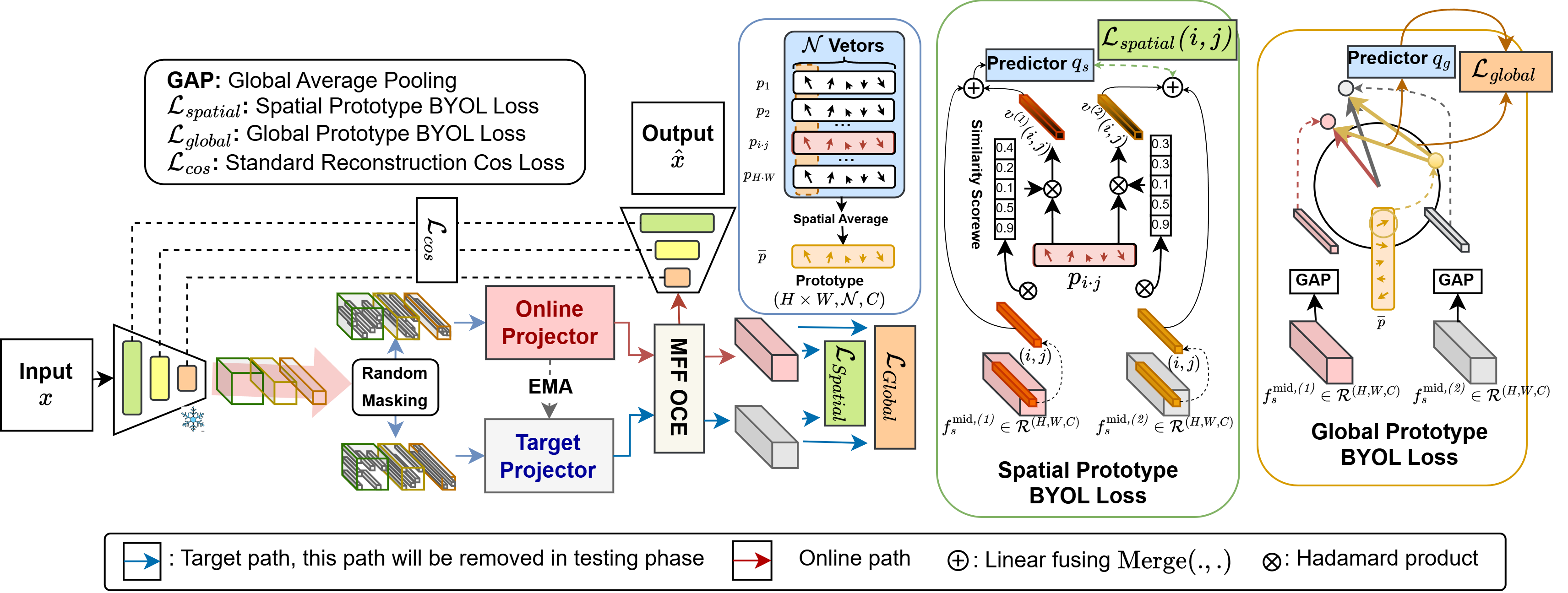}
    \caption{\textbf{Overview of BoRAD.} The framework uses prototype-guided training regularization to allocate representation capacity between local spatial contraction and global discriminability preservation. The reconstruction cosine loss denoted as $\mathcal{L}_{cos}$ in the figure corresponds to $\mathcal{L}_{recon}$ in the text. At inference, anomaly maps are computed from teacher-student feature discrepancy without prototype lookup.}
    \label{fig:overview}
\end{figure}

% \Needspace{18\baselineskip}

BoRAD addresses the representation dilemma in multi-class reconstruction: the student must be capacity-limited enough to reject anomalous identity details, yet expressive enough to keep distinct normal clusters separated.

\begin{wrapfigure}[12]{r}{0.36\textwidth}
    % \vspace{-8pt}
    \centering
    \includegraphics[width=\linewidth]{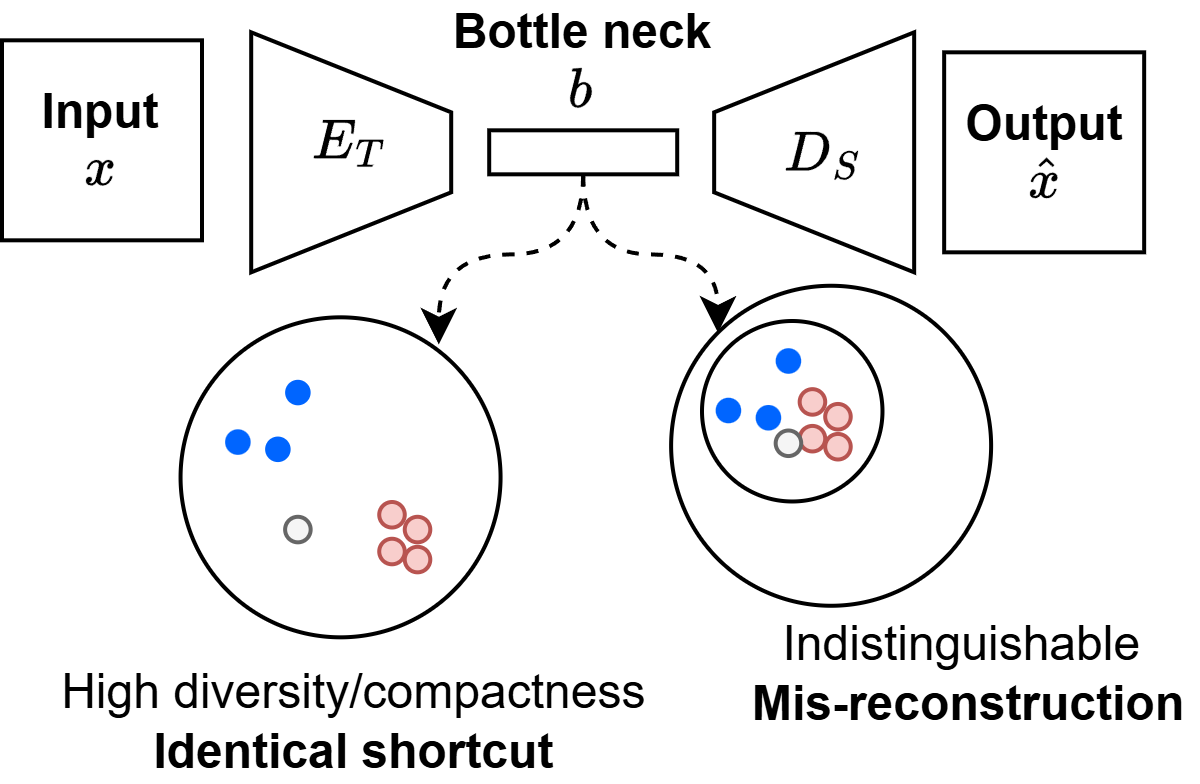}
    \caption{\textbf{The Representation Dilemma.}}
    \label{fig:dilemma}
    % \vspace{-8pt}
\end{wrapfigure}

As shown in Figure~\ref{fig:dilemma}, an over-expressive latent space can pass anomalous features through the decoder, creating an \emph{Identity Shortcut}; an over-compressed space can collapse distinct normal clusters, causing \emph{Inter-class Confusion}. Our goal is to control the latent representation in a balanced way: contract local within-cluster variation while preserving global between-cluster structure. Because over-expressiveness and over-compression reflect excessive and insufficient information retention, respectively, we adopt an information-centric view and relate information to variance; constraining variance through contrastive objectives offers a simple way to guide the latent space away from both extremes.

\paragraph{Setup.}
Given an image $\mathbf{x}$, a frozen teacher $E_T$ extracts dense features $\mathbf{f}_e$. As summarized in Figure~\ref{fig:overview}, the online branch applies a learnable projector $\phi_\theta$ and an MFF-OCE fusion module to obtain an intermediate feature $f^{mid}$, which is decoded by the student $D_S$ into spatial features $\mathbf{f}_s$; its pooled representation is $g=\mathrm{GAP}(f^{mid})$. A momentum target branch $\phi_\xi$ provides stable cross-view targets for augmented views $(1)$ and $(2)$. The reconstruction loss $\mathcal{L}_{recon}$ and the test-time anomaly map are both computed from teacher-student cosine discrepancy, with high local discrepancy indicating anomalous regions.

\subsection{Information Bottleneck View of Shortcut Learning}
\label{sec:ib_shortcut}

Shortcut learning can be viewed as a mismatch between what is sufficient for the training objective and what remains reliable under distribution shift. Let $X$ denote the input, $Y$ the intended target, and $Z=f_\theta(X)$ the learned representation. The Information Bottleneck (IB) principle seeks a representation that preserves task-relevant information while discarding unnecessary input information:
\begin{equation}
    \min_{\theta} \; I(X;Z) - \beta I(Z;Y),
\end{equation}
or equivalently, a representation with minimal input information subject to sufficient predictive power for $Y$~\cite{kawaguchi2023does,tishby2000information}.

In standard shortcut learning, the input can be decomposed into robust task factors $X_c$ and shortcut factors $X_s$. Because $X_s$ can be easier to encode or optimize, a model may preserve $I(Z;X_s)$ while suppressing harder but more reliable factors~\cite{robinson2021can}. Thus, compression alone does not remove shortcuts: an IB objective can still preserve a shortcut if it is the cheapest sufficient statistic for the training distribution.

Reconstruction-based anomaly detection has an analogous failure. The desired representation should preserve normal structure while discarding anomalous variation, but training observes only normal data and never explicitly defines what must be rejected. If the student pathway has excessive capacity, the representation can behave as a near-identity channel:
\begin{equation}
    I(X;Z) \approx \mathcal{H}(X).
\end{equation}
Here, $\mathcal{H}(X)$ denotes the entropy of the input representation, or differential entropy when $X$ is modeled as a continuous feature variable.
At test time, this channel can transmit anomalous components through the bottleneck, allowing the decoder to reconstruct $X_a$ instead of projecting the input back to the normal manifold. We call this failure the \emph{Identity Shortcut}: the model solves reconstruction by preserving input identity rather than learning a normality-constrained representation.

This view clarifies why anomaly detection requires more than ordinary reconstruction accuracy. A useful reconstruction model should be sufficient for normal structure, not for arbitrary input identity. We therefore constrain the representation so that only normal-compatible information remains decodable:
\begin{equation}
    \min \mathcal{L}_{\mathrm{recon}}
    \quad \mathrm{s.t.} \quad
    I(X;Z) \leq \mathcal{C}, \qquad
    Z \in \mathcal{S}_{\mathrm{normal}},
\end{equation}
where $\mathcal{S}_{\mathrm{normal}}$ denotes the learned normal representation manifold. Since $I(X;Z)=h(Z)-h(Z|X)$ and $h(Z|X)$ is constant under our deterministic or fixed-noise encoder view, capacity control reduces to controlling the marginal differential entropy $h(Z)$ up to an additive constant. Directly optimizing $h(Z)$ is impractical, so we use the Maximum Entropy Principle and AM-GM inequality to upper-bound it with a monotonic function of the total variance $\text{Tr}(\Sigma_Z)$ (Appendix~\ref{app:info_capacity}). BoRAD implements this constraint through prototype-guided variance decomposition: local prototype-conditioned alignment contracts within-anchor variation, while global shifted alignment preserves between-anchor variation to avoid collapsing distinct normal clusters.

\subsection{Geometric Bifurcation via the Law of Total Variance}
To prevent the total collapse of the representation space---which would destroy between-cluster discriminability---we partition the Total Variance. Given a set of learnable prototype anchors $\mathcal{P}$, the Law of Total Variance decomposes $\text{Tr}(\Sigma_Z)$ into two orthogonal components:
\begin{equation}
\text{Var}(Z) = \mathbb{E}_{\mathcal{P}}[\text{Var}(Z|\mathcal{P})] + \text{Var}_{\mathcal{P}}(\mathbb{E}[Z|\mathcal{P}])
\end{equation}
This decomposition dictates our architectural topology. We use a geometric bifurcation strategy: local alignment reduces the intra-cluster variance term to form a bottleneck against Identity Shortcuts, while shifted global alignment preserves the inter-cluster variance term needed for semantic separability across diverse normal clusters.

\paragraph{Prototype and shifted-coordinate interpretation.}
BoRAD implements this bifurcation with prototype anchors. Prototype representations can regularize continuous feature spaces by providing learnable basis directions~\cite{lu2023hierarchical}; here, the same prototype bank serves two variance-specific roles. Locally, spatial routing through prototypes targets the conditional term $\mathbb{E}_{\mathcal{P}}[\mathrm{Var}(Z|\mathcal{P})]$: it contracts within-prototype variation and forms a low-rank normal-manifold bottleneck against Identity Shortcuts. Globally, prototype-relative shifting targets the between-prototype term $\mathrm{Var}_{\mathcal{P}}(\mathbb{E}[Z|\mathcal{P}])$: it changes the coordinate system before angular alignment so that normal cluster directions remain separable instead of being uniformly collapsed. This follows the broader intuition behind mean-shifted representation learning~\cite{reiss2023mean}, but differs in construction: BoRAD learns multiple anchors jointly with the reconstruction model rather than shifting by a single fixed mean. The prototypes are not an inference-time memory; they are discarded after training and only shape the teacher-student reconstruction pathway.

\subsection{Spatial Prototype Alignment}
To reduce the expected intra-cluster variance $\mathbb{E}_{\mathcal{P}}[\text{Var}(Z|\mathcal{P})]$, we operate on the local spatial features. However, unconstrained continuous variance reduction can still preserve identical shortcuts for anomalous regions. We therefore introduce a learnable visual dictionary of spatially-indexed prototypes $\mathcal{P} \in \mathbb{R}^{(H \times W) \times \mathcal{N} \times C}$ to provide a low-rank training-time bottleneck. The spatial prototype BYOL block in Figure~\ref{fig:overview} illustrates this per-location prototype routing, Merge fusion, and spatial prediction path.

We force the fusion outputs from both augmented views to query this prototype bank via a cosine-similarity weighted linear projection at each spatial location $(i,j)$:
\begin{equation}
v^{(t)}_{(i,j)} = (f^{mid, (t)}_{s,(i,j)} \mathcal{P}_{(i,j)}) \mathcal{P}_{(i,j)} \quad \text{for} \quad t \in \{1, 2\}
\end{equation}
where $f^{mid,(t)}_{s,(i,j)}$ denotes the student-side intermediate feature of view $t$ at location $(i,j)$, $\mathcal{P}_{(i,j)}$ denotes the corresponding location-specific prototype matrix, and $v^{(t)}_{(i,j)}$ is the prototype-routed feature at that location. After this prototype-guided projection, we enforce a predictive alignment between the two views. The final spatial latent representations are constructed by fusing the raw continuous features with these prototype-conditioned structural summaries: $\mathbf{f}_s^{(t)} = \text{Merge}(f^{mid, (t)}, v^{(i)})$. To directly execute the variance contraction, the online representation $\mathbf{f}_s^{(1)}$ is mapped through a spatial predictor network $q_s$ and aligned against the target representation $\mathbf{f}_s^{(2)}$:
\begin{equation}
\mathcal{L}_{spatial} = 1 - \cos\left(q_s(\mathbf{f}_s^{(1)}), \mathbf{f}_s^{(2)}\right)
\end{equation}
Assuming an optimal spatial predictor, gradient descent on this predictive alignment reduces the conditional variance $\text{Var}(\mathbf{f}_s^{(2)} | \mathbf{f}_s^{(1)})$ (proof in Appendix~\ref{app:predictor_variance}). Because the representation is mediated through the low-rank linear bottleneck of the prototype bank, this variance reduction is biased toward a low-dimensional normal sub-manifold (Appendix~\ref{app:prototype_subspace}). The learned decoder is therefore encouraged to favor normal-manifold directions $\mathcal{S}_{normal} = \text{span}(\mathcal{P})$ over out-of-distribution variations.

\subsection{Global Alignment and Prototype-Relative Angular Sensitivity}
\label{sec:sensitivity}
Spatial alignment contracts local conditional variance, but multi-class reconstruction also needs to preserve the directions that separate normal clusters. We therefore add a global alignment loss in a prototype-relative coordinate system.

Let $\delta_g = \mathbf{g}^{(1)}-\mathbf{g}^{(2)}$ denote the cross-view difference in global feature space. Directly aligning $\mathbf{g}^{(1)}$ and $\mathbf{g}^{(2)}$ measures angles from the global origin, which can blur between-cluster geometry. Instead, we mean-pool the spatial prototype bank into $\mathcal{N}$ normalized global anchors $\tilde{\mathbf{p}}_k$ and align residuals
\begin{equation}
\mathbf{r}_k^{(i)} = \mathbf{g}^{(i)} - \tilde{\mathbf{p}}_k, \qquad \|\tilde{\mathbf{p}}_k\|_2 = 1.
\end{equation}
The resulting BYOL-style global objective is
\begin{equation}
\mathcal{L}_{global} = \frac{1}{\mathcal{N}} \sum_{k=1}^{\mathcal{N}} \left( 1 - \cos\left( q_g(\mathbf{r}_k^{(1)}), \mathbf{r}_k^{(2)} \right) \right).
\end{equation}
Here, $q_g$ is an online predictor. The shift preserves the cross-view difference, $\mathbf{r}_k^{(1)}-\mathbf{r}_k^{(2)}=\delta_g$, but changes the angular coordinate system because cosine similarity is not translation invariant. When a feature direction approaches a prototype direction, the residual becomes
\begin{equation}
\mathbf{r}_k = \|\mathbf{g}\|\tilde{\mathbf{g}} - \tilde{\mathbf{p}}_k
\quad \xrightarrow{\tilde{\mathbf{g}} \to \tilde{\mathbf{p}}_k} \quad
(\|\mathbf{g}\|-1)\tilde{\mathbf{p}}_k
\end{equation}
and retains magnitude information through the energy gap $(\|\mathbf{g}\|-1)$. Intuitively, the prototype shift does not erase the view difference $\delta_g$; it changes where the angle is measured from. A deviation that appears small around the global origin can become more visible when measured relative to nearby prototype anchors. This \emph{prototype-relative angular sensitivity} helps the global loss preserve between-cluster directions while making off-manifold deviations easier to separate, without negative pairs or class labels. Its stationary behavior is analyzed in Appendix~\ref{app:eigenvalue_optimization}.

\subsection{Overall Objective and Synergistic Optimization}
During training, the overall objective of the BoRAD framework is formulated as a weighted sum of the foundational reconstruction, spatial predictive alignment, and global alignment losses shown in Figure~\ref{fig:overview}:
\begin{equation}
\mathcal{L}_{total} = \mathcal{L}_{recon} + \lambda_s \mathcal{L}_{spatial} + \lambda_g \mathcal{L}_{global}
\end{equation}
where $\lambda_s$ and $\lambda_g$ are scalar coefficients dynamically balancing the structural optimization paths. 

The integration of these losses implements the Geometric Bifurcation strategy described in Section 3.2:

\textbf{$\mathcal{L}_{recon}$ (Base Knowledge Distillation):} Establishes the foundational dense mapping from the frozen teacher.

\textbf{$\mathcal{L}_{spatial}$ (Local Variance Contraction):} By routing localized features through the prototype bottleneck and predictive alignment, it reduces intra-cluster conditional variance and discourages the decoder from copying high-frequency anomalous variations.

\textbf{$\mathcal{L}_{global}$ (Between-Cluster Geometry Preservation):} By aligning shifted residuals around the spatial prototype axes, it preserves inter-cluster variance while increasing sensitivity to prototype-relative angular changes. This discourages collapse of diverse semantic representations and reduces Inter-class Confusion in multi-category reconstruction.

Together, these losses reduce anomalous reconstruction capacity while preserving diverse normal cluster structure.

\section{Experimental Results}\label{sec:experimental_results}

\begin{table}[t!]
  \centering
  \caption{Comparison with state-of-the-art methods. Best results are highlighted in bold, and second-best results are underlined.}
  \label{tbl-sota-final}
  \resizebox{\columnwidth}{!}{
  \begin{tabular}{ll ccc cccc c}
    \toprule
    \multirow{2}{*}{Dataset} & \multirow{2}{*}{Method} & \multicolumn{3}{c}{Image} & \multicolumn{4}{c}{Pixel} & \multirow{2}{*}{mAD$\uparrow$} \\
    \cmidrule(lr){3-5} \cmidrule(lr){6-9}
    & & AU-ROC$\uparrow$ & AP$\uparrow$ & F1$\uparrow$ & AU-ROC$\uparrow$ & AP$\uparrow$ & F1$\uparrow$ & AU-PRO$\uparrow$ & \\
    \midrule
    \multirow{10}{*}{MVTec AD} 
    & UniAD \textit{[NeurIPS'22]}~\cite{you2022unified} & 96.5 & 98.8 & 96.2 & 96.8 & 43.4 & 49.5 & 90.7 & 81.7 \\
    & RD4AD \textit{[CVPR'22]}~\cite{deng2022anomaly} & 94.6 & 96.5 & 95.2 & 96.1 & 48.6 & 53.8 & 91.1 & 82.3 \\
    & SimpleNet \textit{[CVPR'23]}~\cite{liu2023simplenet} & 95.3 & 98.4 & 95.8 & 96.9 & 45.9 & 49.7 & 86.5 & 81.2 \\
    & DeSTSeg \textit{[CVPR'23]}~\cite{zhang2023destseg} & 89.2 & 95.5 & 91.6 & 93.1 & 54.3 & 50.9 & 64.8 & 77.1 \\
    & DiAD \textit{[AAAI'24]}~\cite{he2024diffusion} & 97.2 & 99.0 & 96.5 & 96.8 & 52.6 & 55.5 & 90.7 & 84.0 \\
    & HVQ-Trans \textit{[NeurIPS'23]}~\cite{lu2023hierarchical} & 98.0 & 99.5 & 97.5 & 97.3 & 48.2 & 53.3 & 91.4 & 83.6 \\
    & MambaAD \textit{[NeurIPS'24]}~\cite{he2024mambaad} & 98.6 & \underline{99.6} & 97.8 & \underline{97.7} & \textbf{56.3} & \underline{59.2} & 93.1 & \underline{86.0} \\
    & ViTAD \textit{[CVIU'25]}~\cite{zhang2025exploring} & 98.3 & 99.4 & 97.3 & \underline{97.7} & 55.3 & 58.7 & 91.4 & 85.4 \\
    & OmiAD \textit{[ICML'25]}~\cite{feng2025omiad} & \underline{98.8} & \textbf{99.7} & \textbf{98.5} & \underline{97.7} & 52.6 & 56.7 & \underline{93.2} & 85.3 \\
    & \textbf{BoRAD (Ours)} & \textbf{99.1 $\pm$ 0.13} & \underline{99.6 $\pm$ 0.04} & \underline{98.1 $\pm$ 0.16} & \textbf{97.8 $\pm$ 0.04} & \underline{56.0 $\pm$ 0.40} & \textbf{59.4 $\pm$ 0.27} & \textbf{93.8 $\pm$ 0.05} & \textbf{86.2 $\pm$ 0.07} \\
    \midrule
    \multirow{10}{*}{VisA} 
    & UniAD \textit{[NeurIPS'22]}~\cite{you2022unified} & 88.8 & 90.8 & 85.8 & 98.3 & 33.7 & 39.0 & 85.5 & 74.6 \\
    & RD4AD \textit{[CVPR'22]}~\cite{deng2022anomaly} & 92.4 & 92.4 & 89.6 & 98.1 & 38.0 & 42.6 & \underline{91.8} & 77.8 \\
    & SimpleNet \textit{[CVPR'23]}~\cite{liu2023simplenet} & 87.2 & 87.0 & 81.8 & 96.8 & 34.7 & 37.8 & 81.4 & 72.4 \\
    & DeSTSeg \textit{[CVPR'23]}~\cite{zhang2023destseg} & 88.9 & 89.0 & 85.2 & 96.1 & 39.6 & 43.4 & 67.4 & 72.8 \\
    & DiAD \textit{[AAAI'24]}~\cite{he2024diffusion} & 86.8 & 88.3 & 85.1 & 96.0 & 26.1 & 33.0 & 75.2 & 70.1 \\
    & HVQ-Trans \textit{[NeurIPS'23]}~\cite{lu2023hierarchical} & 93.2 & 92.8 & 87.6 & \underline{98.7} & 35.0 & 39.6 & 86.3 & 76.2 \\
    & MambaAD \textit{[NeurIPS'24]}~\cite{he2024mambaad} & 94.3 & 94.5 & 89.4 & 98.5 & 39.4 & 44.0 & 91.0 & 78.7 \\
    & ViTAD \textit{[CVIU'25]}~\cite{zhang2025exploring} & 90.5 & 91.7 & 86.3 & 98.2 & 36.6 & 41.1 & 85.1 & 75.6 \\
    & OmiAD \textit{[ICML'25]}~\cite{feng2025omiad} & \underline{95.3} & \underline{96.0} & \underline{91.2} & \textbf{98.9} & \underline{40.4} & \underline{44.1} & 89.2 & \underline{79.3} \\
    & \textbf{BoRAD (Ours)} & \textbf{95.5 $\pm$ 0.08} & \textbf{96.1 $\pm$ 0.12} & \textbf{91.9 $\pm$ 0.06} & \underline{98.7 $\pm$ 0.03} & \textbf{43.4 $\pm$ 0.18} & \textbf{46.7 $\pm$ 0.15} & \textbf{93.0 $\pm$ 0.13} & \textbf{80.7 $\pm$ 0.02} \\
    \midrule
    \multirow{10}{*}{Real-IAD} 
    & UniAD \textit{[NeurIPS'22]}~\cite{you2022unified} & 83.0 & 80.9 & 74.3 & 97.3 & 21.1 & 29.2 & 86.7 & 67.5 \\
    & RD4AD \textit{[CVPR'22]}~\cite{deng2022anomaly} & 82.4 & 79.0 & 73.9 & 97.3 & 25.0 & 32.7 & 89.6 & 68.6 \\
    & SimpleNet \textit{[CVPR'23]}~\cite{liu2023simplenet} & 57.2 & 53.4 & 61.5 & 75.7 & 2.8 & 6.5 & 39.0 & 42.3 \\
    & DeSTSeg \textit{[CVPR'23]}~\cite{zhang2023destseg} & 82.3 & 79.2 & 73.2 & 94.6 & \textbf{37.9} & \underline{41.7} & 40.6 & 64.2 \\
    & DiAD \textit{[AAAI'24]}~\cite{he2024diffusion} & 75.6 & 66.4 & 69.9 & 88.0 & 2.9 & 7.1 & 58.1 & 52.6 \\
    & HVQ-Trans \textit{[NeurIPS'23]}~\cite{lu2023hierarchical} & 86.6 & 84.9 & \underline{79.4} & 98.0 & 27.6 & 34.4 & 88.7 & 71.4 \\
    & MambaAD \textit{[NeurIPS'24]}~\cite{he2024mambaad} & 86.3 & 84.6 & 77.0 & \underline{98.5} & 33.0 & 38.7 & 90.5 & 72.7 \\
    & ViTAD \textit{[CVIU'25]}~\cite{zhang2025exploring} & -- & -- & -- & -- & -- & -- & -- & -- \\
    & OmiAD \textit{[ICML'25]}~\cite{feng2025omiad} & \textbf{90.1} & \textbf{88.6} & \textbf{82.8} & \textbf{98.9} & \underline{37.7} & \textbf{42.6} & \textbf{93.1} & \textbf{76.3} \\
    & \textbf{BoRAD (Ours)} & \underline{87.4 $\pm$ 0.21} & \underline{85.2 $\pm$ 0.33} & 78.0 $\pm$ 0.16 & 98.2 $\pm$ 0.06 & 32.4 $\pm$ 0.54 & 38.9 $\pm$ 0.38 & \underline{91.3 $\pm$ 0.08} & \underline{73.1 $\pm$ 0.25} \\
    \bottomrule
  \end{tabular}}
\end{table}

In this section, we evaluate BoRAD under a one-for-all setting across three industrial anomaly detection benchmarks. We compare against competitive reconstruction, memory, diffusion, and recent foundation-style anomaly detection baselines, then isolate the contribution of each training-time component through controlled ablations.

\subsection{Experimental Setup}

\textbf{Datasets.} We evaluate on three benchmarks: (1) \textbf{MVTec AD}~\cite{bergmann2019mvtec} (15 industrial categories), (2) \textbf{VisA}~\cite{zou2022spot} (12 categories with complex structures), and (3) \textbf{Real-IAD}~\cite{wang2024real} (large-scale, 30 categories). For each benchmark, a single model is trained on normal images pooled across all categories and evaluated on the corresponding multi-class test split. For the main comparison in Table~\ref{tbl-sota-final} and Table~\ref{tbl-ablation-modules}, we report the mean and standard deviation over three independent runs; for other experimental tables, we report single-run results using seed 42 due to the large number of ablation and diagnostic experiments.

\textbf{Metrics.} We report image-level AU-ROC, AP, and F1-max, together with pixel-level AU-ROC, AP, F1-max, and AU-PRO. \textbf{mAD} is the arithmetic mean of these seven metrics.

\begin{figure*}[t!]
    \centering
    \begin{minipage}[t]{0.52\textwidth}
        \vspace{-0.8em}
        \centering
        \vspace{0pt}
        \includegraphics[width=\linewidth]{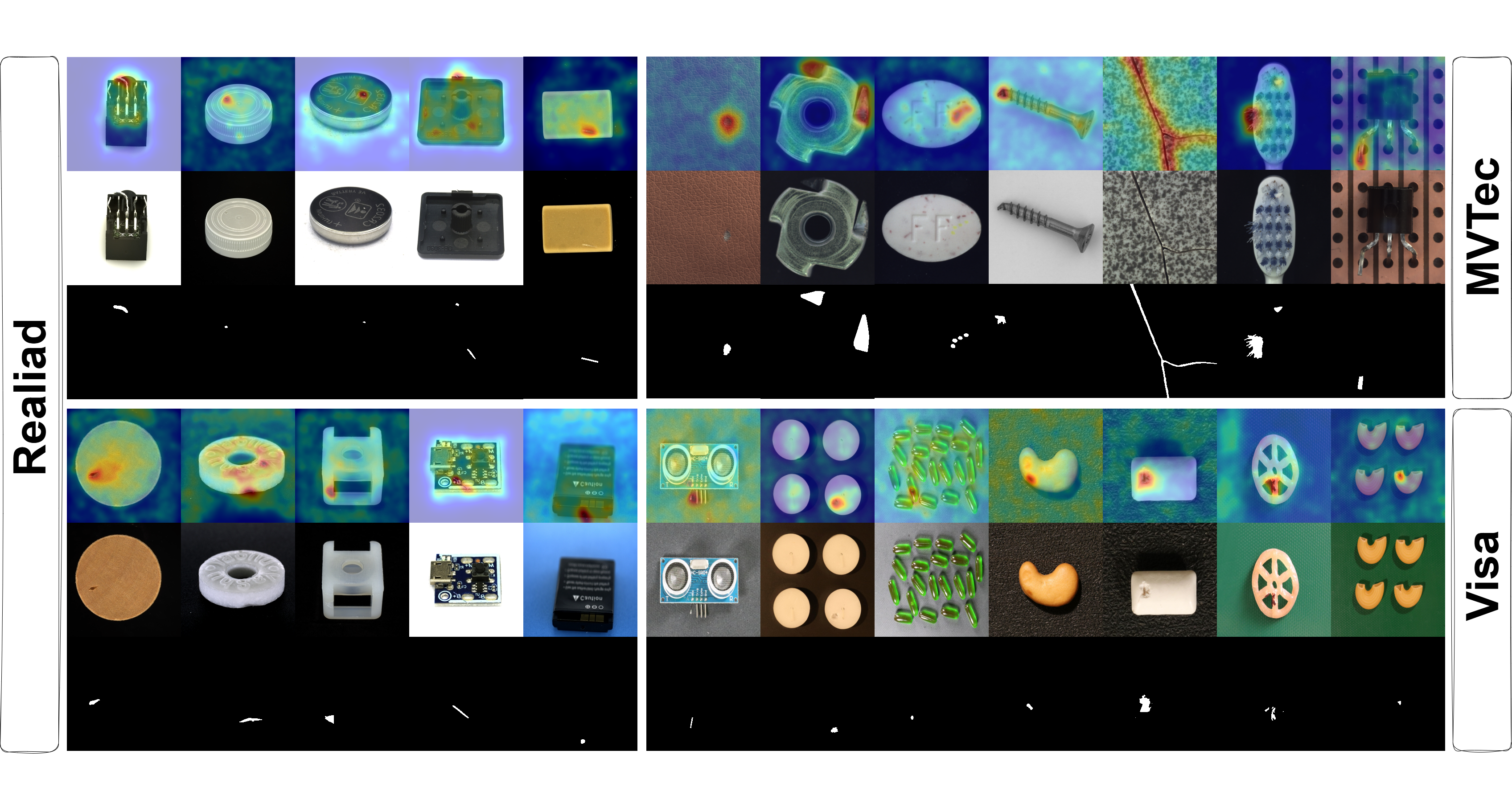}
        \caption{
        \textbf{Qualitative anomaly localization results.}
        Heatmaps, input images, and masks are shown from top to bottom.
        More detail results are in Appendix~\ref{app:qualitative}}
        \label{fig:qualitative_results}
    \end{minipage}
    \hfill
    \begin{minipage}[t]{0.45\textwidth}
        \vspace{-1.0em}
        \centering
        \vspace{0pt}
        \scriptsize
        \setlength{\tabcolsep}{2.5pt}
        \renewcommand{\arraystretch}{0.88}

        \captionof{table}{Ablation on $\lambda_s$.}
        \label{tbl-lambda-spatial}
        \resizebox{\linewidth}{!}{
        \begin{tabular}{lccccccc}
            \toprule
            \multirow{2}{*}{$\lambda_s$} 
            & \multicolumn{3}{c}{Image} 
            & \multicolumn{4}{c}{Pixel} \\
            \cmidrule(lr){2-4} \cmidrule(lr){5-8}
            & AUROC & mAP & mF1 & mAUPRO & mAUROC & mAP & mF1 \\
            \midrule
            0.1 & 98.97 & 99.51 & 97.76 & 93.68 & 97.72 & 55.78 & 59.48 \\
            0.5 & 98.99 & 99.58 & 97.72 & 93.85 & 97.78 & 55.75 & 59.10 \\
            1   & \textbf{99.19} & \textbf{99.63} & \textbf{98.15} & 93.77 & 97.78 & 55.98 & 59.30 \\
            2   & 98.96 & 99.54 & 98.00 & \textbf{93.96} & \textbf{97.82} & \textbf{56.08} & \textbf{59.58} \\
            \bottomrule
        \end{tabular}
        }

        \vspace{0.5em}

        \captionof{table}{Ablation on $\lambda_g$.}
        \label{tbl-lambda-global}
        \resizebox{\linewidth}{!}{
        \begin{tabular}{lccccccc}
            \toprule
            \multirow{2}{*}{$\lambda_g$} 
            & \multicolumn{3}{c}{Image} 
            & \multicolumn{4}{c}{Pixel} \\
            \cmidrule(lr){2-4} \cmidrule(lr){5-8}
            & AUROC & mAP & mF1 & mAUPRO & mAUROC & mAP & mF1 \\
            \midrule
            0.1 & 98.82 & 99.48 & 97.92 & 93.70 & 97.70 & 55.02 & 58.84 \\
            0.5 & 98.96 & 99.51 & 97.73 & 93.76 & 97.73 & 55.21 & 59.05 \\
            1   & \textbf{99.19} & \textbf{99.63} & \textbf{98.15} & 93.77 & 97.78 & \textbf{55.98} & 59.30 \\
            2   & 98.81 & 99.54 & 98.09 & \textbf{93.79} & \textbf{97.85} & 55.93 & \textbf{59.50} \\
            \bottomrule
        \end{tabular}
        }
    \end{minipage}

\end{figure*}

\subsection{Comparison with Competitive Methods}

\begin{wraptable}[9]{r}{0.50\textwidth}
  % \vspace{-12pt}
  \centering
  \caption{\textbf{Parameter and computational efficiency.} FPS evaluated on an NVIDIA RTX 4060 Ti.}
  \label{tbl-efficiency}
  \resizebox{\linewidth}{!}{
  \begin{tabular}{lcccc}
    \toprule
    Mode & Total (M)$\downarrow$ & Trainable (M)$\downarrow$ & GFLOPs$\downarrow$ & FPS$\uparrow$ \\
    \midrule
    Train + proto loss & 120.894 & 80.544 & 52.4 & 62.172 \\
    Test inference & 96.097 & 71.234 & 37.5 & 117.026 \\
    \bottomrule
  \end{tabular}}
  % \vspace{-10pt}
\end{wraptable}

Table~\ref{tbl-sota-final} compares BoRAD with recent one-for-all anomaly detectors, and Table~\ref{tbl-efficiency} reports its parameter and compute cost. BoRAD achieves the best mAD on MVTec AD (\textbf{86.2\%}) and VisA (\textbf{80.7\%}), and the second-best mAD on Real-IAD (\underline{73.1\%}). It is especially strong on MVTec and VisA localization metrics, while OmiAD remains stronger on Real-IAD.

\subsection{Ablation: Validating Components}

\begin{table}[t!]
  \centering
  \caption{Ablation Study on Modules. Best results are highlighted.}
  \label{tbl-ablation-modules}
  \resizebox{\columnwidth}{!}{
  \begin{tabular}{cccc ccc cccc c}
    \toprule
    \multirow{2}{*}{Proto Shift} & \multirow{2}{*}{Masking} & \multirow{2}{*}{Global Loss} & \multirow{2}{*}{Spatial Loss} & \multicolumn{3}{c}{Image} & \multicolumn{4}{c}{Pixel} & \multirow{2}{*}{mAD$\uparrow$} \\
    \cmidrule(lr){5-7} \cmidrule(lr){8-11}
    & & & & AUROC$\uparrow$ & mAP$\uparrow$ & mF1$\uparrow$ & mAUPRO$\uparrow$ & mAUROC$\uparrow$ & mAP$\uparrow$ & mF1$\uparrow$ & \\
    \midrule
    & & & & 98.16 & 99.19 & 96.93 & 93.04 & 97.37 & 53.90 & 57.83 & 85.20 \\
    & \checkmark & \checkmark & \checkmark & 99.01 & 99.52 & 97.91 & 93.65 & 97.70 & 54.36 & 58.66 & 85.83 \\
    \checkmark & & \checkmark & \checkmark & 98.84 & 99.45 & 97.84 & 93.52 & 97.64 & 55.10 & 58.66 & 85.86 \\
    \checkmark & \checkmark & & \checkmark & 98.98 & 99.51 & \textbf{98.10} & 93.76 & 97.76 & 55.25 & 59.04 & 86.06 \\
    \checkmark & \checkmark & \checkmark & & 98.92 & 99.51 & 97.74 & 93.61 & 97.70 & 55.48 & 59.05 & 86.00 \\
    \checkmark & \checkmark & \checkmark & \checkmark & \textbf{99.08} & \textbf{99.59} & 98.05 & \textbf{93.79} & \textbf{97.78} & \textbf{56.01} & \textbf{59.41} & \textbf{86.24} \\
    \bottomrule
  \end{tabular}}
\end{table}

Table~\ref{tbl-ablation-modules} validates the contribution of each BoRAD component on MVTec AD. The base reconstruction model obtains 85.2\% mAD, but its lower pixel AP/F1 indicates limited localization capacity control. Adding the proposed components consistently improves performance. Prototype shifting and feature masking mainly improve pixel-level metrics, suggesting that shifted alignment benefits from stochastic feature perturbations. The global and spatial losses provide complementary gains: the global loss improves overall separability, while the spatial loss achieves good scores in localization. The full model achieves the best results across all metrics, reaching \textbf{99.08\%} image AU-ROC, \textbf{56.01\%} pixel AP, \textbf{59.41\%} pixel F1-max, and 86.2\% mAD. The ablation on the weighting between the spatial and global losses is reported in Table~\ref{tbl-lambda-spatial} and Table~\ref{tbl-lambda-global}. Additional hyperparameter ablations are reported in Appendix~\ref{app:ablations}.

\subsection{Qualitative Analysis}

\begin{figure}[t]
    \centering
    \begin{subfigure}{0.44\textwidth}
        \centering
        \includegraphics[width=\linewidth]{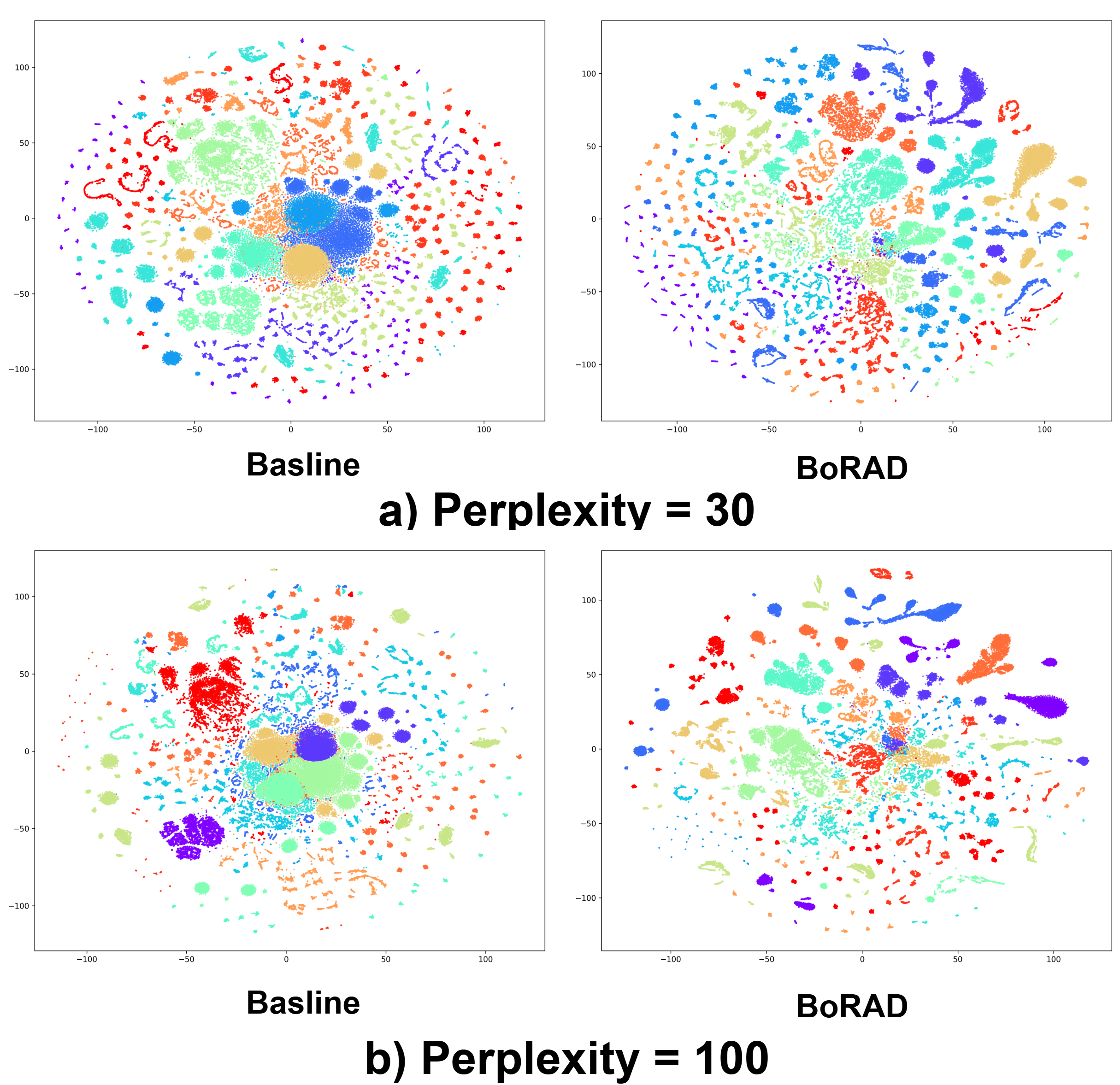}
        \caption{t-SNE~\cite{van2008visualizing} of global features.}
        \label{fig:tsne}
    \end{subfigure}
    \hfill
    \begin{subfigure}{0.54\textwidth}
        \centering
        \includegraphics[width=\linewidth]{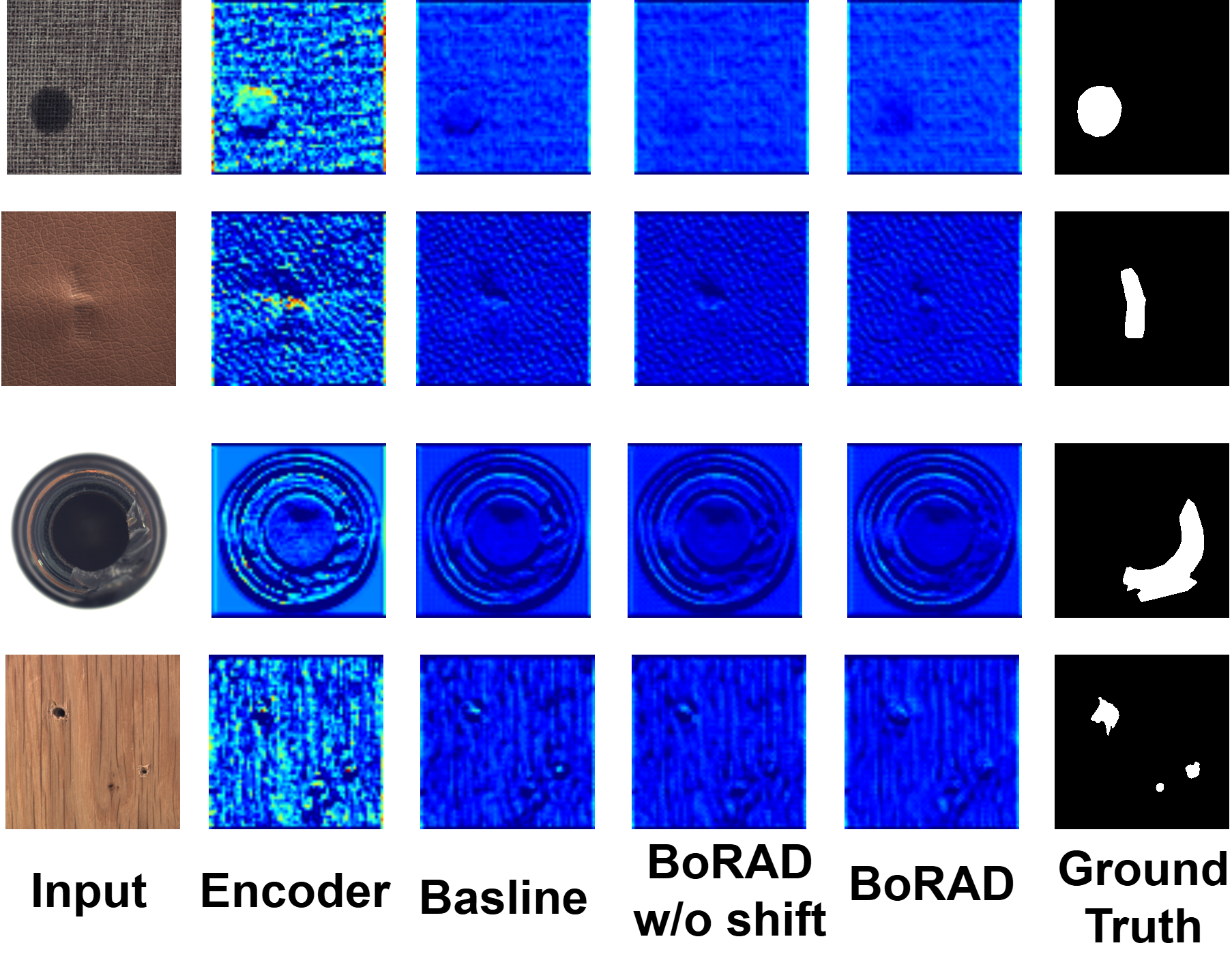}
        \caption{Feature channel activations.}
        \label{fig:shortcut}
    \end{subfigure}
    \caption{\textbf{Qualitative validation of the two training effects.} (a)~t-SNE visualization shows preserved between-cluster separability across categories. (b)~Feature channel activations show local variance contraction: with BoRAD, the student decoder no longer copies the teacher's anomalous activations.}
    \label{fig:combined_horizontal}
\end{figure}

Figure~\ref{fig:combined_horizontal} shows the two expected effects: global features remain category-separated, while local activations no longer copy anomalous teacher responses. Together with the localization examples in Figure~\ref{fig:qualitative_results}, Figure~\ref{fig:combined_horizontal} shows the two expected effects: global features remain category-separated, while local activations no longer copy anomalous teacher responses.

% \subsection{Diagnostic Analysis of the Representation Dilemma}
% \label{sec:diagnostic_analysis}

% To directly test the two failure modes in Figure~\ref{fig:dilemma}, we compute three MVTec AD diagnostics: anomaly leakage/copy score on ground-truth anomalous pixels, class separability on train-normal $g=\mathrm{GAP}(f^{mid})$ features, and anomaly-normal score separation on the final multi-scale anomaly map. Detailed definitions are in Appendix~\ref{app:diagnostic_metrics}.
\subsection{Diagnostic Analysis of the Representation Dilemma}
\label{sec:diagnostic_analysis}

Aggregate AUROC/AP scores do not reveal which part of the representation dilemma is being improved. We therefore keep one focused diagnostic for each question: CopySim/LeakageGap measure anomalous copying, Silhouette measures train-normal category separability, and score Separation/Pixel AP measure whether anomaly maps separate abnormal from normal pixels. Detailed definitions are in Appendix~\ref{app:diagnostic_metrics}.
\begin{table}[ht]
  \centering
  \caption{\textbf{Focused diagnostic measurements on MVTec AD.} Arrows indicate the preferred direction. CopySim and LeakageGap measure anomalous copying, Silhouette measures normal-category separability in $mid$-GAP features, and Separation/Pixel AP measure anomaly-normal localization quality.}
  \label{tbl-diagnostics}
  \resizebox{0.7\textwidth}{!}{
  \begin{tabular}{lccccc}
    \toprule
    Variant & CopySim$_{anom}\downarrow$ & LeakageGap$\downarrow$ & Silhouette$\uparrow$ & Separation$\uparrow$ & Pixel AP$\uparrow$ \\
    \midrule
    Baseline & 0.914 & -0.045 & 0.637 & 0.073 & 0.506 \\
    w/o Shift & 0.905 & -0.052 & 0.696 & 0.087 & 0.520 \\
    w/o Global & 0.903 & -0.053 & 0.718 & 0.089 & 0.529 \\
    Ours & \textbf{0.887} & \textbf{-0.064} & \textbf{0.841} & \textbf{0.104} & \textbf{0.543} \\
    \bottomrule
  \end{tabular}}
\end{table}

Table~\ref{tbl-diagnostics} supports the intended behavior. BoRAD reduces anomalous copying, improves normal-category separability, and gives the strongest anomaly-normal score separation.

\subsection{Empirical Analysis of Feature Statistics}

\begin{wrapfigure}[16]{r}{0.64\textwidth}
    \vspace{-10pt}
    \centering
    \begin{subfigure}{0.48\linewidth}
        \centering
        \includegraphics[width=\linewidth]{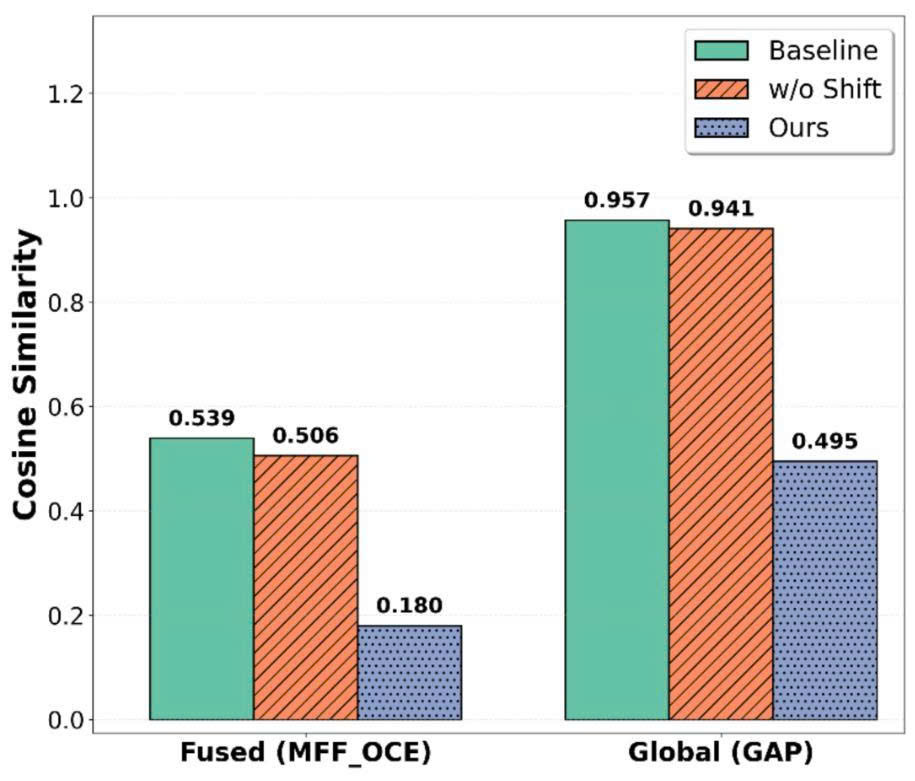}
        \caption{Mean Feature Std}
        \label{fig:feature_std}
    \end{subfigure}
    \hfill
    \begin{subfigure}{0.48\linewidth}
        \centering
        \includegraphics[width=\linewidth]{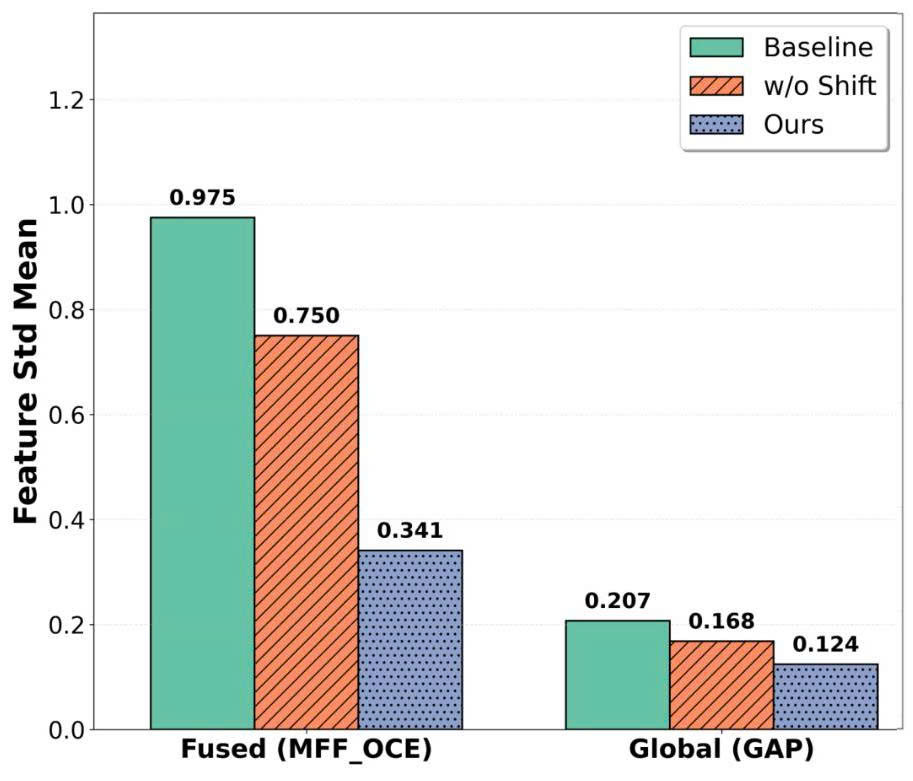}
        \caption{Mean Cosine Sim.}
        \label{fig:cosine_sim}
    \end{subfigure}

    \caption{\textbf{Empirical validation on 100 random test samples.}
    BoRAD with shift reduces feature variation and decreases cross-view cosine similarity.}
    \label{fig:feature_stats}
    % \vspace{-8pt}
\end{wrapfigure}

To quantitatively validate our theoretical claims regarding spatial contraction and angular amplification, we analyze representation statistics on 100 random test samples. We compute the mean feature standard deviation and mean cosine similarity between online and target representations. As shown in Figure~\ref{fig:feature_stats}, BoRAD with shift reduces feature variation, supporting compactness, while decreasing cross-view cosine similarity, indicating stronger angular separation.

\section{Limitations}
\label{sec:limitations}

BoRAD still inherits the inference cost of dense reverse distillation with a high-capacity teacher. Its prototype bottleneck may also be less effective for subtle anomalies that lie close to the learned normal subspace $\mathcal{S}_{normal}$. Finally, the shifted alignment relies on feature masking to induce meaningful cross-view differences; poorly matched perturbations may emphasize sub-optimal directions for some edge-case anomalies.

% =====================================================================

\section{Conclusion} \label{sec:conclusion}

We formulate multi-class anomaly detection as an Information Bottleneck problem governed by a fundamental Representation Dilemma. By analyzing the capacity--separability conflict through the Law of Total Variance, we propose BoRAD, a unified unsupervised framework that addresses this dilemma via Geometric Bifurcation. BoRAD uses a shared learnable visual dictionary: spatial predictive alignment contracts conditional variance through prototype-guided projection, forming a low-rank normal-manifold bottleneck that suppresses identical shortcuts, while global coordinate shift preserves prototype-relative angular structure and avoids the invariance paradox without external supervision or negative pairs. Across MVTec AD, VisA, and Real-IAD, BoRAD achieves highly competitive unsupervised anomaly localization performance. Empirical analyses further validate the complementarity between spatial prototype projection and shifted global alignment. In practice, stronger one-for-all inspection can reduce deployment and manual review costs, but deployment should still include threshold calibration, human review, and monitoring under domain shift to mitigate missed defects or unnecessary rejection.

%%%%%%%%%%%%%%%%%%%%%%%%%%%%%%%%%%%%%%%%%%%%%%%%%%%%%%%%%%%%

\bibliographystyle{unsrt}
\bibliography{references}

@article{grill2020bootstrap,
  title={Bootstrap your own latent-a new approach to self-supervised learning},
  author={Grill, Jean-Bastien and Strub, Florian and Altch{\'e}, Florent and Tallec, Corentin and Richemond, Pierre and Buchatskaya, Elena and Doersch, Carl and Avila Pires, Bernardo and Guo, Zhaohan and Gheshlaghi Azar, Mohammad and others},
  journal={Advances in neural information processing systems},
  volume={33},
  pages={21271--21284},
  year={2020}
}

@inproceedings{zong2018deep,
  title={Deep autoencoding gaussian mixture model for unsupervised anomaly detection},
  author={Zong, Bo and Song, Qi and Min, Martin Renqiang and Cheng, Wei and Lumezanu, Cristian and Cho, Daeki and Chen, Haifeng},
  booktitle={International conference on learning representations},
  year={2018}
}

@article{schlegl2019f,
  title={f-AnoGAN: Fast unsupervised anomaly detection with generative adversarial networks},
  author={Schlegl, Thomas and Seeb{\"o}ck, Philipp and Waldstein, Sebastian M and Langs, Georg and Schmidt-Erfurth, Ursula},
  journal={Medical image analysis},
  volume={54},
  pages={30--44},
  year={2019},
  publisher={Elsevier}
}

@inproceedings{li2024promptad,
  title={Promptad: Zero-shot anomaly detection using text prompts},
  author={Li, Yiting and Goodge, Adam and Liu, Fayao and Foo, Chuan-Sheng},
  booktitle={Proceedings of the IEEE/CVF winter conference on applications of computer vision},
  pages={1093--1102},
  year={2024}
}

@inproceedings{lv2025one,
  title={One-for-all few-shot anomaly detection via instance-induced prompt learning},
  author={Lv, Wenxi and Su, Qinliang and Xu, Wenchao},
  booktitle={The Thirteenth International Conference on Learning Representations},
  year={2025}
}

@inproceedings{chen2024unified,
  title={A unified anomaly synthesis strategy with gradient ascent for industrial anomaly detection and localization},
  author={Chen, Qiyu and Luo, Huiyuan and Lv, Chengkan and Zhang, Zhengtao},
  booktitle={European Conference on Computer Vision},
  pages={37--54},
  year={2024},
  organization={Springer}
}

@inproceedings{gong2019memorizing,
  title={Memorizing normality to detect anomaly: Memory-augmented deep autoencoder for unsupervised anomaly detection},
  author={Gong, Dong and Liu, Lingqiao and Le, Vuong and Saha, Budhaditya and Mansour, Moussa Reda and Venkatesh, Svetha and Hengel, Anton van den},
  booktitle={Proceedings of the IEEE/CVF international conference on computer vision},
  pages={1705--1714},
  year={2019}
}

@inproceedings{roth2022towards,
  title={Towards total recall in industrial anomaly detection},
  author={Roth, Karsten and Pemula, Latha and Zepeda, Joaquin and Sch{\"o}lkopf, Bernhard and Brox, Thomas and Gehler, Peter},
  booktitle={Proceedings of the IEEE/CVF conference on computer vision and pattern recognition},
  pages={14318--14328},
  year={2022}
}

@inproceedings{gudovskiy2022cflow,
  title={Cflow-ad: Real-time unsupervised anomaly detection with localization via conditional normalizing flows},
  author={Gudovskiy, Denis and Ishizaka, Shun and Kozuka, Kazuki},
  booktitle={Proceedings of the IEEE/CVF winter conference on applications of computer vision},
  pages={98--107},
  year={2022}
}

@inproceedings{lu2026maskad,
  title={MaskAD: Parallel Masked Autoencoder for Multi-class Unsupervised Anomaly Detection},
  author={Lu, Ruiying and Liu, Gang and Li, Kang and Tian, Long and Zhang, Junwei},
  booktitle={Proceedings of the AAAI Conference on Artificial Intelligence},
  volume={40},
  number={18},
  pages={15457--15465},
  year={2026}
}

@inproceedings{you2022adtr,
  title={Adtr: Anomaly detection transformer with feature reconstruction},
  author={You, Zhiyuan and Yang, Kai and Luo, Wenhan and Cui, Lei and Zheng, Yu and Le, Xinyi},
  booktitle={International Conference on Neural Information Processing},
  pages={298--310},
  year={2022},
  organization={Springer}
}

@article{ulyanov2016instance,
  title={Instance normalization: The missing ingredient for fast stylization},
  author={Ulyanov, Dmitry and Vedaldi, Andrea and Lempitsky, Victor},
  journal={arXiv preprint arXiv:1607.08022},
  year={2016}
}

@inproceedings{yao2023one,
  title={One-for-all: Proposal masked cross-class anomaly detection},
  author={Yao, Xincheng and Zhang, Chongyang and Li, Ruoqi and Sun, Jun and Liu, Zhenyu},
  booktitle={Proceedings of the AAAI Conference on Artificial Intelligence},
  volume={37},
  number={4},
  pages={4792--4800},
  year={2023}
}

@article{bergmann2018improving,
  title={Improving unsupervised defect segmentation by applying structural similarity to autoencoders},
  author={Bergmann, Paul and L{\"o}we, Sindy and Fauser, Michael and Sattlegger, David and Steger, Carsten},
  journal={arXiv preprint arXiv:1807.02011},
  year={2018}
}

@article{guo2023recontrast,
  title={Recontrast: Domain-specific anomaly detection via contrastive reconstruction},
  author={Guo, Jia and Lu, Shuai and Jia, Lize and Zhang, Weihang and Li, Huiqi},
  journal={Advances in Neural Information Processing Systems},
  volume={36},
  pages={10721--10740},
  year={2023}
}

@inproceedings{deng2009imagenet,
  title={Imagenet: A large-scale hierarchical image database},
  author={Deng, Jia and Dong, Wei and Socher, Richard and Li, Li-Jia and Li, Kai and Fei-Fei, Li},
  booktitle={2009 IEEE conference on computer vision and pattern recognition},
  pages={248--255},
  year={2009},
  organization={Ieee}
}

@inproceedings{park2020learning,
  title={Learning memory-guided normality for anomaly detection},
  author={Park, Hyunjong and Noh, Jongyoun and Ham, Bumsub},
  booktitle={Proceedings of the IEEE/CVF conference on computer vision and pattern recognition},
  pages={14372--14381},
  year={2020}
}

@inproceedings{zavrtanik2021draem,
  title={Draem-a discriminatively trained reconstruction embedding for surface anomaly detection},
  author={Zavrtanik, Vitjan and Kristan, Matej and Sko{\v{c}}aj, Danijel},
  booktitle={Proceedings of the IEEE/CVF international conference on computer vision},
  pages={8330--8339},
  year={2021}
}

@article{yang2020dfr,
  title={Dfr: Deep feature reconstruction for unsupervised anomaly segmentation},
  author={Yang, Jie and Shi, Yong and Qi, Zhiquan},
  journal={arXiv preprint arXiv:2012.07122},
  year={2020}
}

@inproceedings{gonzalez2025hyperbolic,
  title={Is Hyperbolic Space All You Need for Medical Anomaly Detection?},
  author={Gonzalez-Jimenez, Alvaro and Lionetti, Simone and Amruthalingam, Ludovic and Gottfrois, Philippe and Gr{\"o}ger, Fabian and Pouly, Marc and Navarini, Alexander A},
  booktitle={International Conference on Medical Image Computing and Computer-Assisted Intervention},
  pages={312--322},
  year={2025},
  organization={Springer}
}

@article{han2021madgan,
  title={MADGAN: Unsupervised medical anomaly detection GAN using multiple adjacent brain MRI slice reconstruction},
  author={Han, Changhee and Rundo, Leonardo and Murao, Kohei and Noguchi, Tomoyuki and Shimahara, Yuki and Milacski, Zolt{\'a}n {\'A}d{\'a}m and Koshino, Saori and Sala, Evis and Nakayama, Hideki and Satoh, Shin’ichi},
  journal={BMC bioinformatics},
  volume={22},
  number={Suppl 2},
  pages={31},
  year={2021},
  publisher={Springer}
}

@inproceedings{maas2013rectifier,
  title={Rectifier nonlinearities improve neural network acoustic models},
  author={Maas, Andrew L and Hannun, Awni Y and Ng, Andrew Y and others},
  booktitle={Proc. icml},
  volume={30},
  number={1},
  pages={3},
  year={2013},
  organization={Atlanta, GA}
}

@article{van2008visualizing,
  title={Visualizing data using t-SNE.},
  author={Van der Maaten, Laurens and Hinton, Geoffrey},
  journal={Journal of machine learning research},
  volume={9},
  number={11},
  year={2008}
}

@article{tishby2000information,
  title={The information bottleneck method},
  author={Tishby, Naftali and Pereira, Fernando C and Bialek, William},
  journal={arXiv preprint physics/0004057},
  year={2000}
}

@inproceedings{saranrittichai2022overcoming,
  title={Overcoming shortcut learning in a target domain by generalizing basic visual factors from a source domain},
  author={Saranrittichai, Piyapat and Mummadi, Chaithanya Kumar and Blaiotta, Claudia and Munoz, Mauricio and Fischer, Volker},
  booktitle={European Conference on Computer Vision},
  pages={294--309},
  year={2022},
  organization={Springer}
}

@article{bengio2013representation,
  title={Representation learning: A review and new perspectives},
  author={Bengio, Yoshua and Courville, Aaron and Vincent, Pascal},
  journal={IEEE transactions on pattern analysis and machine intelligence},
  volume={35},
  number={8},
  pages={1798--1828},
  year={2013},
  publisher={IEEE}
}

@inproceedings{deng2022anomaly,
  title={Anomaly detection via reverse distillation from one-class embedding},
  author={Deng, Hanqiu and Li, Xingyu},
  booktitle={Proceedings of the IEEE/CVF conference on computer vision and pattern recognition},
  pages={9737--9746},
  year={2022}
}

@inproceedings{feng2025omiad,
  title={OmiAD: One-step adaptive masked diffusion model for multi-class anomaly detection via adversarial distillation},
  author={Feng, Yaoxuan and Chen, Wenchao and Li, Yuxin and Chen, Bo and Wang, Yubiao and Zhao, Zixuan and Liu, Hongwei and Zhou, Mingyuan},
  booktitle={Forty-second International Conference on Machine Learning},
  year={2025}
}

@inproceedings{liu2025unlocking,
  title={Unlocking the potential of reverse distillation for anomaly detection},
  author={Liu, Xinyue and Wang, Jianyuan and Leng, Biao and Zhang, Shuo},
  booktitle={Proceedings of the AAAI Conference on Artificial Intelligence},
  volume={39},
  number={6},
  pages={5640--5648},
  year={2025}
}

@article{liu2024deep,
  title={Deep industrial image anomaly detection: A survey},
  author={Liu, Jiaqi and Xie, Guoyang and Wang, Jinbao and Li, Shangnian and Wang, Chengjie and Zheng, Feng and Jin, Yaochu},
  journal={Machine Intelligence Research},
  volume={21},
  number={1},
  pages={104--135},
  year={2024}
}

@inproceedings{you2022unified,
  title={A Unified Model for Multi-class Anomaly Detection},
  author={You, Zhiyuan and Cui, Lei and Shen, Yujun and Yang, Kai and Lu, Xin and Zheng, Yu and Le, Xinyi},
  booktitle={Advances in Neural Information Processing Systems},
  year={2022}
}

@inproceedings{bergmann2019mvtec,
  title={MVTec AD--A comprehensive real-world dataset for unsupervised anomaly detection},
  author={Bergmann, Paul and Fauser, Michael and Sattlegger, David and Steger, Carsten},
  booktitle={IEEE Conference on Computer Vision and Pattern Recognition},
  year={2019}
}

@inproceedings{jeong2023winclip,
  title={Winclip: Zero-/few-shot anomaly classification and segmentation},
  author={Jeong, Jongheon and Zou, Yang and Kim, Taewan and Zhang, Dongqing and Ravichandran, Avinash and Dabeer, Onkar},
  booktitle={Proceedings of the IEEE/CVF conference on computer vision and pattern recognition},
  pages={19606--19616},
  year={2023}
}

@article{lu2023hierarchical,
  title={Hierarchical vector quantized transformer for multi-class unsupervised anomaly detection},
  author={Lu, Ruiying and Wu, YuJie and Tian, Long and Wang, Dongsheng and Chen, Bo and Liu, Xiyang and Hu, Ruimin},
  journal={Advances in Neural Information Processing Systems},
  volume={36},
  pages={8487--8500},
  year={2023}
}

@article{he2024mambaad,
  title={Mambaad: Exploring state space models for multi-class unsupervised anomaly detection},
  author={He, Haoyang and Bai, Yuhu and Zhang, Jiangning and He, Qingdong and Chen, Hongxu and Gan, Zhenye and Wang, Chengjie and Li, Xiangtai and Tian, Guanzhong and Xie, Lei},
  journal={Advances in Neural Information Processing Systems},
  volume={37},
  pages={71162--71187},
  year={2024}
}

@inproceedings{fan2025salvaging,
  title={Salvaging the Overlooked: Leveraging Class-Aware Contrastive Learning for Multi-Class Anomaly Detection},
  author={Fan, Lei and Huang, Junjie and Di, Donglin and Su, Anyang and Song, Tianyou and Pagnucco, Maurice and Song, Yang},
  booktitle={Proceedings of the IEEE/CVF International Conference on Computer Vision},
  pages={21419--21428},
  year={2025}
}

@article{oord2018representation,
  title={Representation learning with contrastive predictive coding},
  author={Oord, Aaron van den and Li, Yazhe and Vinyals, Oriol},
  journal={arXiv preprint arXiv:1807.03748},
  year={2018}
}

@inproceedings{liu2023simplenet,
  title={Simplenet: A simple network for image anomaly detection and localization},
  author={Liu, Zhikang and Zhou, Yiming and Xu, Yuansheng and Wang, Zilei},
  booktitle={Proceedings of the IEEE/CVF conference on computer vision and pattern recognition},
  pages={20402--20411},
  year={2023}
}

@inproceedings{zhang2023destseg,
  title={Destseg: Segmentation guided denoising student-teacher for anomaly detection},
  author={Zhang, Xuan and Li, Shiyu and Li, Xi and Huang, Ping and Shan, Jiulong and Chen, Ting},
  booktitle={Proceedings of the IEEE/CVF conference on computer vision and pattern recognition},
  pages={3914--3923},
  year={2023}
}

@inproceedings{he2024diffusion,
  title={A diffusion-based framework for multi-class anomaly detection},
  author={He, Haoyang and Zhang, Jiangning and Chen, Hongxu and Chen, Xuhai and Li, Zhishan and Chen, Xu and Wang, Yabiao and Wang, Chengjie and Xie, Lei},
  booktitle={Proceedings of the AAAI conference on artificial intelligence},
  volume={38},
  number={8},
  pages={8472--8480},
  year={2024}
}

@inproceedings{wang2021dense,
  title={Dense contrastive learning for self-supervised visual pre-training},
  author={Wang, Xinlong and Zhang, Rufeng and Shen, Chunhua and Kong, Tao and Li, Lei},
  booktitle={Proceedings of the IEEE/CVF conference on computer vision and pattern recognition},
  pages={3024--3033},
  year={2021}
}

@inproceedings{reiss2023mean,
  title={Mean-shifted contrastive loss for anomaly detection},
  author={Reiss, Tal and Hoshen, Yedid},
  booktitle={Proceedings of the AAAI conference on artificial intelligence},
  volume={37},
  number={2},
  pages={2155--2162},
  year={2023}
}

@inproceedings{gu2024rethinking,
  title={Rethinking reverse distillation for multi-modal anomaly detection},
  author={Gu, Zhihao and Zhang, Jiangning and Liu, Liang and Chen, Xu and Peng, Jinlong and Gan, Zhenye and Jiang, Guannan and Shu, Annan and Wang, Yabiao and Ma, Lizhuang},
  booktitle={Proceedings of the AAAI Conference on Artificial Intelligence},
  volume={38},
  number={8},
  pages={8445--8453},
  year={2024}
}

@article{kingma2014adam,
  title={Adam: A method for stochastic optimization},
  author={Kingma, Diederik P and Ba, Jimmy},
  journal={arXiv preprint arXiv:1412.6980},
  year={2014}
}

@article{zagoruyko2016wide,
  title={Wide residual networks},
  author={Zagoruyko, Sergey and Komodakis, Nikos},
  journal={arXiv preprint arXiv:1605.07146},
  year={2016}
}

@inproceedings{zou2022spot,
  title={Spot-the-difference self-supervised pre-training for anomaly detection and segmentation},
  author={Zou, Yang and Jeong, Jongheon and Pemula, Latha and Zhang, Dongqing and Dabeer, Onkar},
  booktitle={European conference on computer vision},
  pages={392--408},
  year={2022},
  organization={Springer}
}

@inproceedings{wang2024real,
  title={Real-iad: A real-world multi-view dataset for benchmarking versatile industrial anomaly detection},
  author={Wang, Chengjie and Zhu, Wenbing and Gao, Bin-Bin and Gan, Zhenye and Zhang, Jiangning and Gu, Zhihao and Qian, Shuguang and Chen, Mingang and Ma, Lizhuang},
  booktitle={Proceedings of the IEEE/CVF Conference on Computer Vision and Pattern Recognition},
  pages={22883--22892},
  year={2024}
}

@article{robinson2021can,
  title={Can contrastive learning avoid shortcut solutions?},
  author={Robinson, Joshua and Sun, Li and Yu, Ke and Batmanghelich, Kayhan and Jegelka, Stefanie and Sra, Suvrit},
  journal={Advances in neural information processing systems},
  volume={34},
  pages={4974--4986},
  year={2021}
}

@inproceedings{kawaguchi2023does,
  title={How does information bottleneck help deep learning?},
  author={Kawaguchi, Kenji and Deng, Zhun and Ji, Xu and Huang, Jiaoyang},
  booktitle={International conference on machine learning},
  pages={16049--16096},
  year={2023},
  organization={PMLR}
}

@article{zhang2025exploring,
  title={Exploring plain vit features for multi-class unsupervised visual anomaly detection},
  author={Zhang, Jiangning and Chen, Xuhai and Wang, Yabiao and Wang, Chengjie and Liu, Yong and Li, Xiangtai and Yang, Ming-Hsuan and Tao, Dacheng},
  journal={Computer Vision and Image Understanding},
  volume={253},
  pages={104308},
  year={2025},
  publisher={Elsevier}
}

% \newpage
% \input{checklist.tex}

% =====================================================================
% APPENDIX
% =====================================================================
\newpage
\appendix
\setcounter{figure}{0}
\renewcommand{\thefigure}{A.\arabic{figure}}
\section{Inference Pseudocode}
\label{app:pseudocode}

Algorithm~\ref{alg:borad_inference} presents the inference pipeline of BoRAD. All training-time components---the prototype bank $\mathcal{P}$, merge layer $\Phi_{\mathrm{merge}}$, predictors ($\Phi_{\mathrm{pred}}$, $\Phi_{\mathrm{gpred}}$), and the momentum target branch---are discarded after training. These components serve as training regularizers that shape the projector and decoder weights during optimization. At test time, inference reduces to a single efficient forward pass: the frozen teacher features are compared against the student's reconstruction via per-scale cosine distance.

\begin{algorithm}[htpb]
\caption{BoRAD Inference}
\label{alg:borad_inference}
\begin{algorithmic}[1]
\State \textbf{Input:} Test image $\mathbf{x}$; frozen encoder $E$; trained projector $\phi_\theta$, MFF-OCE, decoder $S_\omega$
\State \textbf{Discarded:} Prototypes $\mathcal{P}$, predictors $\Phi_{\mathrm{pred}}, \Phi_{\mathrm{gpred}}$, merge $\Phi_{\mathrm{merge}}$, target branch $\phi_\xi$
\State \textbf{Output:} Anomaly map $\mathcal{M}$, image-level score $s_{\text{image}}$
\vspace{1.5mm}
\State \textit{\% 1. Feature extraction}
\State $\{\mathbf{f}_e^{(l)}\}_{l=1}^{3} \gets E(\mathbf{x})$ \Comment{Multi-scale teacher features (frozen)}
\State $\{\mathbf{f}_p^{(l)}\}_{l=1}^{3} \gets \phi_\theta(\{\mathbf{f}_e^{(l)}\})$ \Comment{Online projection}
\State $\mathbf{f}_{\text{fused}} \gets \text{MFF-OCE}(\{\mathbf{f}_p^{(l)}\})$ \Comment{Multi-scale fusion}
\State $\{\mathbf{f}_s^{(l)}\}_{l=1}^{3} \gets S_\omega(\mathbf{f}_{\text{fused}})$ \Comment{Decoder reconstruction}
\vspace{1.5mm}
\State \textit{\% 2. Anomaly scoring}
\State $\mathcal{M}^{(l)} \gets 1 - \cos(\mathbf{f}_e^{(l)},\; \mathbf{f}_s^{(l)})$ \textbf{for each} scale $l$ \Comment{Per-scale cosine distance}
\State $\mathcal{M} \gets \text{Upsample}\!\left(\textstyle\sum_l \mathcal{M}^{(l)}\right)$ \Comment{Fused anomaly map}
\State $s_{\text{image}} \gets \max(\mathcal{M})$
\vspace{1.5mm}
\State \Return $\mathcal{M},\; s_{\text{image}}$
\end{algorithmic}
\end{algorithm}

\section{Mathematical Proofs and Theoretical Derivations}
This appendix provides mathematical derivations for the theoretical framework presented in the main manuscript, covering the optimization dynamics, geometric constraints, and spectral properties of our Reverse Distillation with Prototype-based Coordinate Shifts architecture.

\subsection{Information Capacity Bound via Total Variance}
\label{app:info_capacity}
\textbf{Objective:} Prove that minimizing the Total Variance $\text{Tr}(\Sigma_Z)$ provides a monotonic upper bound on the Information Capacity $I(X; Z)$.

\textbf{Proof:} Let $X$ be the input data and $Z = E_S(X)$ be the latent representation extracted by the student network. The Information Capacity of the latent space is defined by the Mutual Information:
\begin{equation}
I(X; Z) = h(Z) - h(Z|X)
\end{equation}
In a deterministic implementation, $Z|X$ is degenerate; under the common fixed-noise stochastic encoder interpretation, $h(Z|X)$ is finite but constant because the conditional covariance is fixed. In both cases, this conditional term is not optimized toward zero by the reconstruction objective. We therefore control $I(X;Z)$ through the marginal differential entropy $h(Z)$ up to an additive constant.

By the Maximum Entropy Principle, for a random variable $Z \in \mathbb{R}^d$ with a known covariance matrix $\Sigma_Z$, the differential entropy is strictly upper-bounded by that of a multivariate Gaussian distribution $\mathcal{N}(\mu, \Sigma_Z)$:
\begin{equation}
h(Z) \le \frac{1}{2} \log \left( (2\pi e)^d \det(\Sigma_Z) \right) = \frac{d}{2} \log(2\pi e) + \frac{1}{2} \log(\det(\Sigma_Z))
\end{equation}
Because $\Sigma_Z$ is a symmetric and positive semi-definite covariance matrix, its determinant is the product of its eigenvalues $\lambda_i$:
\begin{equation}
\det(\Sigma_Z) = \prod_{i=1}^d \lambda_i
\end{equation}
To avoid the computational instability of optimizing the log-determinant in high-dimensional deep networks, we apply the Arithmetic Mean-Geometric Mean (AM-GM) inequality. The geometric mean of non-negative real numbers is strictly bounded by their arithmetic mean:
\begin{equation}
\left( \prod_{i=1}^d \lambda_i \right)^{1/d} \le \frac{1}{d} \sum_{i=1}^d \lambda_i
\end{equation}
Recognizing that the sum of the eigenvalues equals the trace of the matrix ($\sum_{i=1}^d \lambda_i = \text{Tr}(\Sigma_Z)$), we rewrite the inequality as:
\begin{equation}
\det(\Sigma_Z)^{1/d} \le \frac{1}{d} \text{Tr}(\Sigma_Z)
\end{equation}
Taking the natural logarithm of both sides and multiplying by $d$:
\begin{equation}
\log(\det(\Sigma_Z)) \le d \log \left( \frac{1}{d} \text{Tr}(\Sigma_Z) \right)
\end{equation}
Substituting this back into the entropy bound yields:
\begin{equation}
h(Z) \le \frac{d}{2} \log(2\pi e) + \frac{d}{2} \log \left( \frac{\text{Tr}(\Sigma_Z)}{d} \right)
\end{equation}
Because the logarithm is a monotonically increasing function, minimizing the trace $\text{Tr}(\Sigma_Z)$ strictly minimizes the upper bound of the latent space's differential entropy, effectively compressing the representational volume. 

\subsection{Predictor Optimality and Conditional Variance}
\label{app:predictor_variance}
\textbf{Objective:} Prove that the gradient updates of the predictive alignment minimize the expected conditional variance of the target given the online representation.

\textbf{Proof:} Let $z^{(1)}$ and $z^{(2)}$ be two augmented views of the same spatial representation. The alignment minimizes the Mean Squared Error (MSE) using a predictor network $q_\phi$ parameterized by $\phi$, while the online encoder is parameterized by $\theta$. The loss objective is:
\begin{equation}
\mathcal{L}(\theta, \phi) = \mathbb{E}_{X, \text{aug}} \left[ \|q_\phi(f^{(1)}_\theta) - f^{(2)}\|_2^2 \right]
\end{equation}
Assuming the neural predictor $q_\phi$ has sufficient capacity and is optimized to its theoretical limit $q^*$, the optimal estimator for minimizing MSE is mathematically defined as the conditional expectation:
\begin{equation}
q^*(f^{(1)}_\theta) = \mathbb{E} \left[ f^{(2)} | f^{(1)}_\theta \right]
\end{equation}
Substituting this optimal predictor formulation back into the loss function, the objective for updating the online network parameters $\theta$ becomes:
\begin{equation}
\mathcal{L}(\theta) = \mathbb{E}_{X, \text{aug}} \left[ \left\| \mathbb{E} \left[ f^{(2)} | f^{(1)}_\theta \right] - f^{(2)} \right\|_2^2 \right]
\end{equation}
By the statistical definition of variance, the expected squared $L_2$ distance between a random vector and its conditional expectation is exactly the sum of the conditional variances of its components:
\begin{equation}
\mathbb{E} \left[ \| \mathbb{E}[Y|X] - Y \|_2^2 \right] = \sum_i \text{Var}(Y_i | X)
\end{equation}
Therefore, the loss function simplifies to the expected conditional variance:
\begin{equation}
\mathcal{L}(\theta) = \mathbb{E} \left[ \sum_i \text{Var} \left( f^{(2)}_i | f^{(1)}_\theta \right) \right]
\end{equation}
Gradient descent on $\theta$ minimizes this conditional-variance objective under the optimal-predictor approximation. It encourages the online representation $f^{(1)}_\theta$ to act as a localized contraction mapping, stripping away high-frequency augmentation noise to reliably predict the smoothed centroid. 

\subsection{Prototype-Guided Linear Subspace Projection}
\label{app:prototype_subspace}
\textbf{Objective:} Show that routing the feature through the prototype bank $\mathcal{P}$ via linear projection restricts the continuous variance reduction to a low-dimensional sub-manifold, acting as an orthogonal rejection filter for anomalies.

\textbf{Proof:} Let $f^{(1)} \in \mathbb{R}^C$ be the continuous online feature vector at a specific spatial location. The prototype bank is defined as a matrix $\mathcal{P} \in \mathbb{R}^{\mathcal{N} \times C}$, containing $\mathcal{N}$ learnable basis vectors. The activation-free spatial routing mechanism calculates the similarity scores and reconstructs the feature in a single linear operation:
\begin{equation}
v^{(1)} = (f^{(1)} \mathcal{P}^T) \mathcal{P}
\end{equation}
Let $S = f^{(1)} \mathcal{P}^T \in \mathbb{R}^{1 \times \mathcal{N}}$ be the linear projection coefficients. The routed feature $v^{(1)}$ can be explicitly rewritten as a linear combination of the prototype row vectors $p_k$:
\begin{equation}
v^{(1)} = \sum_{k=1}^{\mathcal{N}} s_k p_k
\end{equation}
Mathematically, this places the routed output in the linear span (the column space of $\mathcal{P}^T$) of the prototype bank:
\begin{equation}
v^{(1)} \in \mathcal{S}_{normal} = \text{span}(p_1, p_2, \dots, p_K)
\end{equation}
Assuming $\mathcal{N} < C$, $\mathcal{S}_{normal}$ forms a low-dimensional linear subspace within the original feature space $\mathbb{R}^C$. Applying the optimal predictor principle (from Appendix~\ref{app:predictor_variance}), the gradient dynamics minimize the conditional variance subject to this geometric bottleneck:
\begin{equation}
\nabla_\theta \mathcal{L}_{spatial} \propto \nabla_\theta \mathbb{E}_{X} \left[ \text{Var} \left( f^{(2)} | \mathcal{S}_{normal}, f^{(1)} \right) \right]
\end{equation}
This imposes a low-rank structural limit. By the Orthogonal Decomposition Theorem, any input feature $f^{(1)}$ can be decomposed into a parallel component $f_{\parallel} \in \mathcal{S}_{normal}$ and an orthogonal component $f_{\perp} \in \mathcal{S}_{normal}^{\perp}$. When the learned dictionary is close to a useful normal basis, the routing $(f^{(1)} \mathcal{P}^T) \mathcal{P}$ favors components representable by $\mathcal{S}_{normal}$ over components outside that span.

Because $\mathcal{P}$ is optimized on normal data, it is encouraged to span the principal directions of normal variation. If an out-of-distribution anomaly lies outside or weakly overlaps this span, the routing mechanism behaves as a low-rank rejection filter: it attenuates anomalous components that cannot be represented by the learned dictionary. This reduces $\mathbb{E}_{\mathcal{P}}[\text{Var}(Z|\mathcal{P})]$ on normal data and establishes the intended dictionary bottleneck.

\subsection{Implicit Eigenvalue Optimization}
\label{app:eigenvalue_optimization}
\textbf{Objective:} Prove that minimizing the cosine distance of augmented residuals implicitly solves an eigenvalue problem for the covariance matrix $\Sigma$ of cross-view feature differences.

\textbf{Proof:} Let $r=\frac{1}{2}(r^{(1)}+r^{(2)})$ be the midpoint residual of two global views, and let $\delta=r^{(1)}-r^{(2)}$ be their cross-view feature difference with covariance $\mathbb{E}[\delta \delta^T] = \Sigma$. Equivalently, the two residual views can be written as $r^{(1)} = r + \frac{\delta}{2}$ and $r^{(2)} = r - \frac{\delta}{2}$. The objective is to minimize the expected cosine distance:
\begin{equation}
\mathcal{L}_{global}(r) = \mathbb{E}_{\delta} \left[ 1 - \frac{r^{(1)T} r^{(2)}}{\|r^{(1)}\| \|r^{(2)}\|} \right]
\end{equation}
1. Numerator Expansion:
\begin{equation}
r^{(1)T} r^{(2)} = \left(r + \frac{\delta}{2}\right)^T \left(r - \frac{\delta}{2}\right) = \|r\|^2 - \frac{1}{4}\|\delta\|^2
\end{equation}
2. Denominator Taylor Approximation:
Assuming the cross-view difference is small ($\|\delta\| \ll \|r\|$), we expand the product of the norms:
\begin{equation}
\|r^{(1)}\| \|r^{(2)}\| = \sqrt{\left(\|r\|^2 + \frac{1}{4}\|\delta\|^2\right)^2 - (r^T \delta)^2}
\end{equation}
Using the first-order Taylor expansion $\sqrt{A^2 - B} \approx A - \frac{B}{2A}$:
\begin{equation}
\|r^{(1)}\| \|r^{(2)}\| \approx \|r\|^2 + \frac{1}{4}\|\delta\|^2 - \frac{1}{2} \frac{(r^T \delta)^2}{\|r\|^2}
\end{equation}
3. Fractional Simplification:
Applying the approximation $\frac{A-\epsilon}{A+\gamma} \approx 1 - \frac{\epsilon+\gamma}{A}$ for small $\epsilon, \gamma$:
\begin{equation}
\frac{\|r\|^2 - \frac{1}{4}\|\delta\|^2}{\|r\|^2 + \frac{1}{4}\|\delta\|^2 - \frac{1}{2}\frac{(r^T \delta)^2}{\|r\|^2}} \approx 1 - \frac{\|\delta\|^2}{2\|r\|^2} + \frac{(r^T \delta)^2}{2\|r\|^4}
\end{equation}
Thus, the cosine loss simplifies to:
\begin{equation}
\mathcal{L}_{global}(r) \approx \mathbb{E}_{\delta} \left[ \frac{\|\delta\|^2}{2\|r\|^2} - \frac{(r^T \delta)^2}{2\|r\|^4} \right]
\end{equation}
4. Expectation over Cross-View Differences:
Applying the expectation operator, noting that $\mathbb{E}[\|\delta\|^2] = \text{Tr}(\Sigma)$ and $\mathbb{E}[(r^T \delta)^2] = \mathbb{E}[r^T \delta \delta^T r] = r^T \Sigma r$:
\begin{equation}
\mathcal{L}_{global}(r) \approx \frac{1}{2} \left( \frac{\text{Tr}(\Sigma)}{\|r\|^2} - \frac{r^T \Sigma r}{\|r\|^4} \right)
\end{equation}
5. Gradient and Stationary Condition:
Taking the derivative with respect to $r$ using standard vector calculus identities ($\nabla_r (r^T \Sigma r) = 2\Sigma r$ and $\nabla_r (\|r\|^{-n}) = -n\|r\|^{-(n+2)}r$):
\begin{equation}
\nabla_r \mathcal{L}_{global} = \frac{1}{2} \left[ \frac{-2 \text{Tr}(\Sigma)r}{\|r\|^4} - \left( \frac{2 \Sigma r}{\|r\|^4} - \frac{4 (r^T \Sigma r)r}{\|r\|^6} \right) \right]
\end{equation}
Setting the stationary condition $\nabla_r \mathcal{L}_{global} = 0$:
\begin{equation}
\frac{-2 \text{Tr}(\Sigma)r}{\|r\|^4} - \frac{2 \Sigma r}{\|r\|^4} + \frac{4 (r^T \Sigma r)r}{\|r\|^6} = 0
\end{equation}
Multiplying the entire equation by $-\frac{1}{2} \|r\|^4$ yields:
\begin{equation}
\text{Tr}(\Sigma)r + \Sigma r - 2 \frac{r^T \Sigma r}{\|r\|^2} r = 0
\end{equation}
Rearranging to isolate the covariance projection on $r$:
\begin{equation}
\Sigma r = \left( 2 \frac{r^T \Sigma r}{\|r\|^2} - \text{Tr}(\Sigma) \right) r
\end{equation}
The term within the parentheses evaluates to a data-dependent scalar value. Denoting this scalar as $\lambda$, the optimization trajectory rigorously resolves to:
\begin{equation}
\Sigma r = \lambda r
\end{equation}
This confirms that gradient descent on the normalized angular distance in the shifted residual space inherently forces the residual vector $r$ to align with eigenvectors of the local cross-view feature-difference covariance $\Sigma$.

\subsection{Low-Rank Dictionary Bottleneck vs. Rigid Projection}
\label{sec:appendix_low_rank}
The spatial alignment mechanism employs the routing $v^{(1)} = (f^{(1)} \mathcal{P}^T) \mathcal{P}$, where $\mathcal{P} \in \mathbb{R}^{\mathcal{N} \times C}$ denotes the prototype bank. During network initialization, the rows of $\mathcal{P}$ are seeded as strictly orthonormal basis vectors using Singular Value Decomposition (SVD). However, during continuous end-to-end optimization via backpropagation, $\mathcal{P}$ is parameterized as an unconstrained parameter matrix without an explicit orthogonal regularization penalty (i.e., we do not definitively enforce $\mathcal{P}\mathcal{P}^T = I$).

Consequently, this operation is not a rigid orthogonal projection in the strict mathematical sense. Instead, it functions as an unconstrained low-rank linear bottleneck, analogous to a sparse dictionary lookup. Because the latent dictionary capacity is restricted ($\mathcal{N} \ll C$) and $\mathcal{P}$ is optimized to reduce conditional variance on the normal data manifold, the gradient dynamics encourage the sub-manifold $\mathcal{S}_{normal} = \text{span}(\mathcal{P})$ to capture principal variations of the normal training density.

When out-of-distribution variations (anomalies) emerge during inference, they may occupy high-frequency sub-spaces orthogonal or weakly aligned with $\mathcal{S}_{normal}$. The routing mechanism acts as a rejection filter not because of an explicitly forced orthogonality constraint, but due to the limited capacity of the learned basis $\mathcal{P}$ to represent patterns absent from the normal training distribution. 

\subsection{Model Architecture and Stochastic Augmentation Details}
\label{sec:appendix_architecture}
\textbf{Spatial Predictor Architecture}: The spatial query-and-predict pipeline consists of two sub-modules inside the \texttt{PrototypeBYOLLoss} module. The \textbf{Merge} head is a single-layer $1 \times 1$ ConvNet that takes the concatenated $(2C)$-channel feature and maps it back to $C$ channels: $1\times1\text{ Conv} \rightarrow \texttt{InstanceNorm2d} \rightarrow \texttt{LeakyReLU}$. The \textbf{Spatial Predictor} $q_s$ is a two-layer MLP built with $1 \times 1$ convolutions:
\begin{align*}
q_s:\;& \texttt{Conv2d(1x1)} \to \texttt{InstanceNorm2d} \to \texttt{LeakyReLU} \\
& \to \texttt{Conv2d(1x1)} \to \texttt{InstanceNorm2d} \to \texttt{LeakyReLU}.
\end{align*}
The use of instance-level normalizations keeps spatial patch statistics localized without intermingling them with macroscopic mini-batch statistics. The \textbf{Global Predictor} $q_g$ is a symmetric two-layer MLP operating on $C$-dimensional vectors: $\texttt{Linear}(C, C) \to \texttt{InstanceNorm1d} \to \texttt{LeakyReLU} \to \texttt{Linear}(C, C) \to \texttt{InstanceNorm1d} \to \texttt{LeakyReLU}$.

\textbf{Feature-level Bernoulli Masking Generator}: In contrast to conventionally augmenting anomalous patterns directly within the input image RGB domain, our formulation applies perturbations within the raw teacher feature space, \emph{before} any student-side processing. Given the three intermediate feature maps $\{f_1, f_2, f_3\}$ extracted from the frozen Wide-ResNet-50-2 teacher at output indices $\{1, 2, 3\}$, two distinct symmetric views are generated independently at each scale using a per-spatial-location Bernoulli dropout tensor:
\begin{align}
\text{Mask}_1^{(i)} &\sim \text{Bernoulli}(1 - p_{mask}) \\
\text{Mask}_2^{(i)} &\sim \text{Bernoulli}(1 - p_{mask})
\end{align}
where $p_{mask} = 0.5$ and $i \in \{1,2,3\}$ indexes each scale. The views are derived via element-wise spatial multiplication: $f_i^{(1)} = f_i \odot \text{Mask}_1^{(i)}$ and $f_i^{(2)} = f_i \odot \text{Mask}_2^{(i)}$. This masking is applied \emph{stochastically per training iteration}: with probability 0.5, the full clean features are used for both views (i.e., $f_i^{(1)} = f_i^{(2)} = f_i$) to prevent over-regularization during early training. The masked features are then passed independently through the MFF-OCE module to produce the two fused views $f^{mid}_{\text{online}}$ and $f^{mid}_{\text{target}}$. This masking strategy induces the cross-view feature difference $\delta$, whose covariance $\Sigma$ drives the shifted cosine alignment discussed in Theorem 1.

% =====================================================================
\section{Implementation Details}
\label{app:implementation}

\textbf{Architecture.} Wide-ResNet50-2~\cite{zagoruyko2016wide} pre-trained on ImageNet serves as the frozen encoder, extracting features from stages 2, 3, and 4. The decoder mirrors the encoder with transposed convolutions. All projection and prediction layers use Instance Normalization~\cite{ulyanov2016instance} and Leaky ReLU~\cite{maas2013rectifier}. 

\begin{table}[h]
  \centering
  \caption{\textbf{Parameter breakdown by module.} The teacher and EMA target branch are included in total parameters but are frozen during training.}
  \label{tbl-param-breakdown}
  \resizebox{\columnwidth}{!}{
  \begin{tabular}{lccp{0.50\columnwidth}}
    \toprule
    Module & Total (M)$\downarrow$ & Trainable (M)$\downarrow$ & Role / reason \\
    \midrule
    net\_t teacher WRN & 24.863 & 0.000 & Frozen teacher in \texttt{train()} mode. \\
    proj\_layer & 15.487 & 15.487 & Trainable projection layer. \\
    proj\_layer\_momentum & 15.487 & 0.000 & EMA target branch; \texttt{requires\_grad=False}. \\
    mff\_oce & 30.830 & 30.830 & Trainable multi-scale fusion and OCE module. \\
    net\_s decoder & 24.918 & 24.918 & Trainable student decoder. \\
    PrototypeBYOLLoss & 9.310 & 9.310 & Trainable prototype and prediction losses. \\
    \bottomrule
  \end{tabular}}
\end{table}

\textbf{Optimizers.} We use separate Adam~\cite{kingma2014adam} optimizers: (i)~\texttt{proj\_opt} for the predictor, (ii)~\texttt{distill\_opt} for the projector and decoder, and (iii)~\texttt{proto\_opt} for prototypes, merge layer, and prototype predictors. All use $\beta_1=0.5$, $\beta_2=0.999$, lr$=5\times10^{-3}$ with step decay $\times 0.1$ at epoch~80.

\textbf{Data augmentation.} Image-level preprocessing uses resize and ImageNet normalization. The two training views used by the prototype alignment losses are generated in feature space using the stochastic Bernoulli masking described in Appendix~\ref{sec:appendix_architecture}; no RGB-level anomaly synthesis is used.

% =====================================================================
\section{Diagnostic Metric Definitions}
\label{app:diagnostic_metrics}

This appendix specifies the diagnostic measurements used in Table~\ref{tbl-diagnostics}. These measurements are not used for model selection or training; they are post-hoc analyses designed to test whether BoRAD reduces anomalous copying, preserves normal-category separability, and improves anomaly-normal score separation.

\textbf{Identity Shortcut: anomaly leakage and copy score.} We evaluate teacher-student copying on anomalous MVTec AD test images with ground-truth masks. For each feature scale $s$, let $f_{t,s}\in\mathbb{R}^{C_s\times H_s\times W_s}$ denote the teacher feature and $f_{r,s}\in\mathbb{R}^{C_s\times H_s\times W_s}$ denote the student reconstruction. At each spatial location $(x,y)$, we compute
\begin{equation}
\mathrm{CopySim}_s(x,y)=
\cos\!\left(f_{t,s}[:,x,y], f_{r,s}[:,x,y]\right).
\end{equation}
The binary ground-truth anomaly mask is resized to $(H_s,W_s)$ with nearest-neighbor interpolation. We then average CopySim over anomalous and normal locations:
\begin{align}
\mathrm{CopySim}_{anom,s} &= \mathbb{E}_{(x,y):M_s(x,y)=1}\left[\mathrm{CopySim}_s(x,y)\right],\\
\mathrm{CopySim}_{norm,s} &= \mathbb{E}_{(x,y):M_s(x,y)=0}\left[\mathrm{CopySim}_s(x,y)\right].
\end{align}
The reported leakage diagnostics are averaged over feature scales and images:
\begin{equation}
\mathrm{LeakageGap} = \mathrm{CopySim}_{anom}-\mathrm{CopySim}_{norm}.
\end{equation}
A lower anomalous CopySim indicates weaker teacher-student copying on abnormal pixels. A more negative LeakageGap indicates that anomalous regions are copied less faithfully than normal regions.

\textbf{Inter-class Confusion: normal-category separability.} We evaluate separability on train-normal samples. For each image, we use the fused projected feature $f^{mid}$ and compute the global representation
\begin{equation}
g = \frac{\mathrm{GAP}(f^{mid})}{\|\mathrm{GAP}(f^{mid})\|_2}.
\end{equation}
Given category labels, we report the cosine silhouette score:
\begin{equation}
\mathrm{Silhouette}=\mathrm{silhouette\_score}\left(\{g_i\},\{c_i\},\mathrm{metric}=\mathrm{cosine}\right),
\end{equation}
where a higher value indicates clearer category-level clustering among normal samples.

\textbf{Prototype-relative angular sensitivity: anomaly-normal score separation.} We compute the final anomaly score map from multi-scale teacher-student discrepancies. At scale $s$,
\begin{equation}
\mathrm{score}_s(x,y)=1-\cos\!\left(f_{t,s}[:,x,y], f_{r,s}[:,x,y]\right).
\end{equation}
Each score map is resized to the mask or image resolution with bilinear interpolation, and the final score is averaged across the $S$ feature scales:
\begin{equation}
\mathrm{ScoreMap}=\frac{1}{S}\sum_{s=1}^{S}\mathrm{Resize}\left(\mathrm{score}_s\right).
\end{equation}
Using the nearest-neighbor resized ground-truth mask $M$, we compute
\begin{align}
\mathrm{Score}_{anom} &= \mathbb{E}_{(x,y):M(x,y)=1}\left[\mathrm{ScoreMap}(x,y)\right],\\
\mathrm{Score}_{norm} &= \mathbb{E}_{(x,y):M(x,y)=0}\left[\mathrm{ScoreMap}(x,y)\right],\\
\mathrm{Separation} &= \mathrm{Score}_{anom}-\mathrm{Score}_{norm}.
\end{align}
Pixel AP is computed from the same score map and mask labels. To control memory and class imbalance, we sample at most 8192 pixels per anomalous image and use the corresponding ScoreMap values as anomaly scores.

% =====================================================================
\section{Hyperparameter Ablation Studies}
\label{app:ablations}

All ablations are conducted on MVTec AD with the default configuration unless otherwise stated.

\begin{table}[h!]
  \centering
  \caption{Ablation on number of prototypes $\mathcal{N}$.}
  \label{tbl-ablation}
  \begin{tabular}{lccccccc}
    \toprule
    \multirow{2}{*}{$\mathcal{N}$} & \multicolumn{3}{c}{Image} & \multicolumn{4}{c}{Pixel} \\
    \cmidrule(lr){2-4} \cmidrule(lr){5-8}
    & AUROC$\uparrow$ & mAP$\uparrow$ & mF1$\uparrow$ & mAUPRO$\uparrow$ & mAUROC$\uparrow$ & mAP$\uparrow$ & mF1$\uparrow$ \\
    \midrule
    3  & 98.96 & 99.53 & \textbf{98.16} & \textbf{93.96} & \textbf{97.80} & \textbf{56.05} & \textbf{59.54} \\
    5  & 98.98 & 99.55 & 98.06 & 93.86 & 97.77 & 55.96 & 59.20 \\
    7  & \textbf{99.19} & \textbf{99.63} & 98.15 & 93.77 & 97.78 & 55.98 & 59.30 \\
    10 & 98.77 & 99.47 & 97.97 & 93.83 & 97.76 & 55.47 & 59.12 \\
    \bottomrule
  \end{tabular}
\end{table}

\begin{table}[h]
  \centering
  \caption{Ablation on momentum schedule.}
  \label{tbl-scheduler}
  \begin{tabular}{llccccccc}
    \toprule
    \multirow{2}{*}{Type} & \multirow{2}{*}{Schedule} & \multicolumn{3}{c}{Image} & \multicolumn{4}{c}{Pixel} \\
    \cmidrule(lr){3-5} \cmidrule(lr){6-9}
    & & AUROC$\uparrow$ & mAP$\uparrow$ & mF1$\uparrow$ & mAUPRO$\uparrow$ & mAUROC$\uparrow$ & mAP$\uparrow$ & mF1$\uparrow$ \\
    \midrule
    \multirow{3}{*}{Constant} & 0.9 & 98.92 & 99.46 & 97.74 & 93.69 & 97.70 & 55.27 & 58.77 \\
    & 0.99 & 99.07 & 99.63 & 98.15 & \textbf{93.95} & \textbf{97.84} & \textbf{56.56} & \textbf{59.61} \\
    & 0.999 & 98.99 & 99.55 & 97.92 & 93.81 & 97.80 & 55.63 & 59.00 \\
    \midrule
    Linear & 0.99--0.999 & \textbf{99.19} & \textbf{99.63} & \textbf{98.15} & 93.77 & 97.78 & 55.98 & 59.30 \\
    \bottomrule
  \end{tabular}
\end{table}

\begin{table}[h!]
  \centering
  \caption{Ablation on masking ratio $\rho$.}
  \label{tbl-mask-ratio}
  \begin{tabular}{lccccccc}
    \toprule
    \multirow{2}{*}{$\rho$} & \multicolumn{3}{c}{Image} & \multicolumn{4}{c}{Pixel} \\
    \cmidrule(lr){2-4} \cmidrule(lr){5-8}
    & AUROC$\uparrow$ & mAP$\uparrow$ & mF1$\uparrow$ & mAUPRO$\uparrow$ & mAUROC$\uparrow$ & mAP$\uparrow$ & mF1$\uparrow$ \\
    \midrule
    0.1 & 98.91 & 99.47 & 97.94 & 93.68 & 97.73 & 55.45 & 58.97 \\
    0.3 & 99.15 & \textbf{99.65} & 98.13 & 93.88 & 97.73 & 55.87 & 59.26 \\
    0.5 & \textbf{99.19} & 99.63 & \textbf{98.15} & 93.77 & 97.78 & 55.98 & 59.30 \\
    0.7 & 99.14 & 99.64 & 98.06 & \textbf{94.03} & \textbf{97.83} & \textbf{56.32} & \textbf{59.97} \\
    0.9 & 98.67 & 99.47 & 97.94 & 93.92 & 97.74 & 55.75 & 59.41 \\
    \bottomrule
  \end{tabular}
\end{table}

\section{Additional Qualitative Results}
\label{app:qualitative}

In this section, we provide extended qualitative visualizations of anomaly localization to further demonstrate the robustness and precision of BoRAD across different challenging benchmarks. For all visualizations, the top row displays our predicted anomaly score map (heatmap) overlaid on the input, the middle row shows the original anomalous input image, and the bottom row presents the ground-truth binary mask of the defect.

\begin{figure}[h!]
    \centering
    \includegraphics[width=0.8\textwidth]{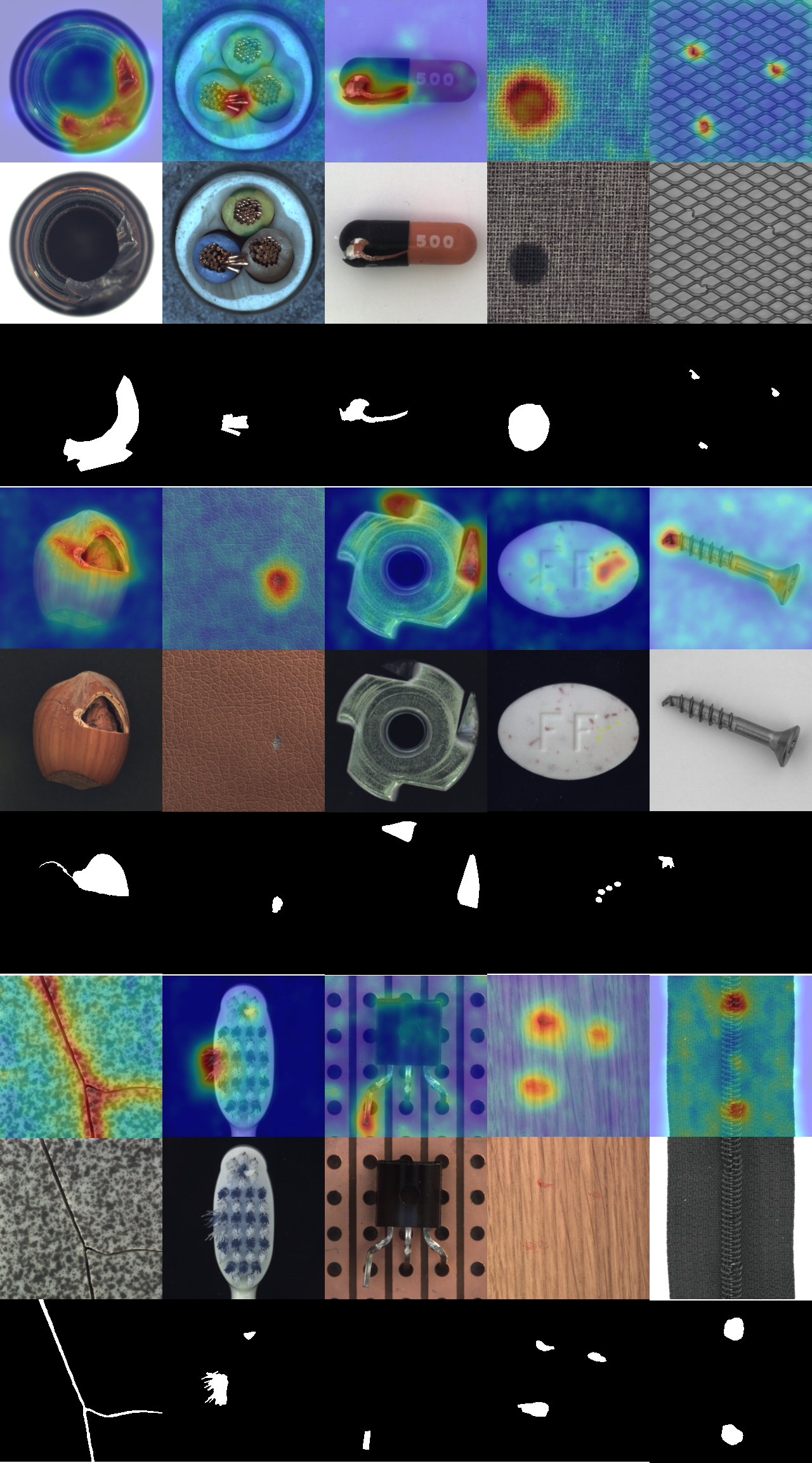}
    \caption{\textbf{Qualitative results on the MVTec AD dataset.} Top row: Predicted anomaly maps. Middle row: Original anomalous images. Bottom row: Ground-truth masks.}
    \label{fig:quali_mvtec}
\end{figure}

\begin{figure}[h!]
    \centering
    \includegraphics[width=\textwidth]{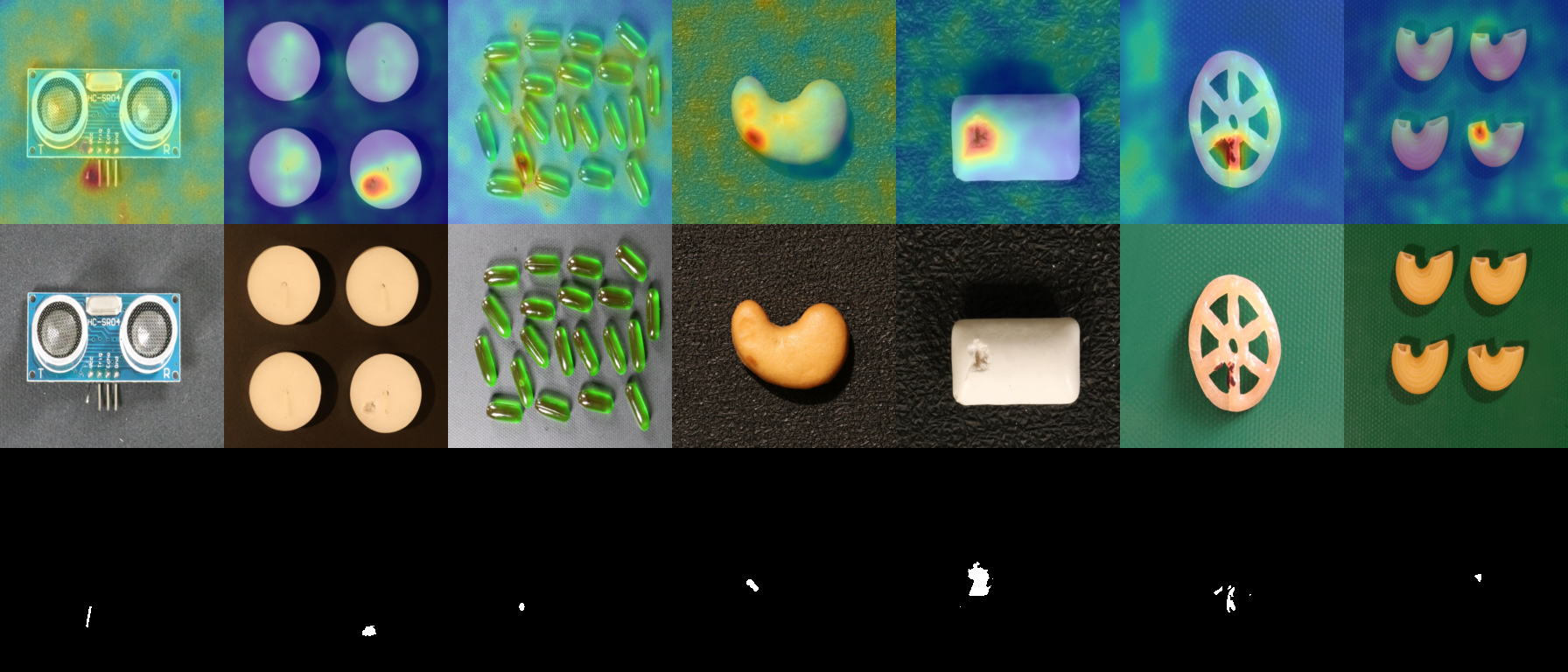}
    \caption{\textbf{Qualitative results on the VisA dataset.} Top row: Predicted anomaly maps. Middle row: Original anomalous images. Bottom row: Ground-truth masks.}
    \label{fig:quali_visa}
\end{figure}

\begin{figure}[h!]
    \centering
    \includegraphics[width=\textwidth]{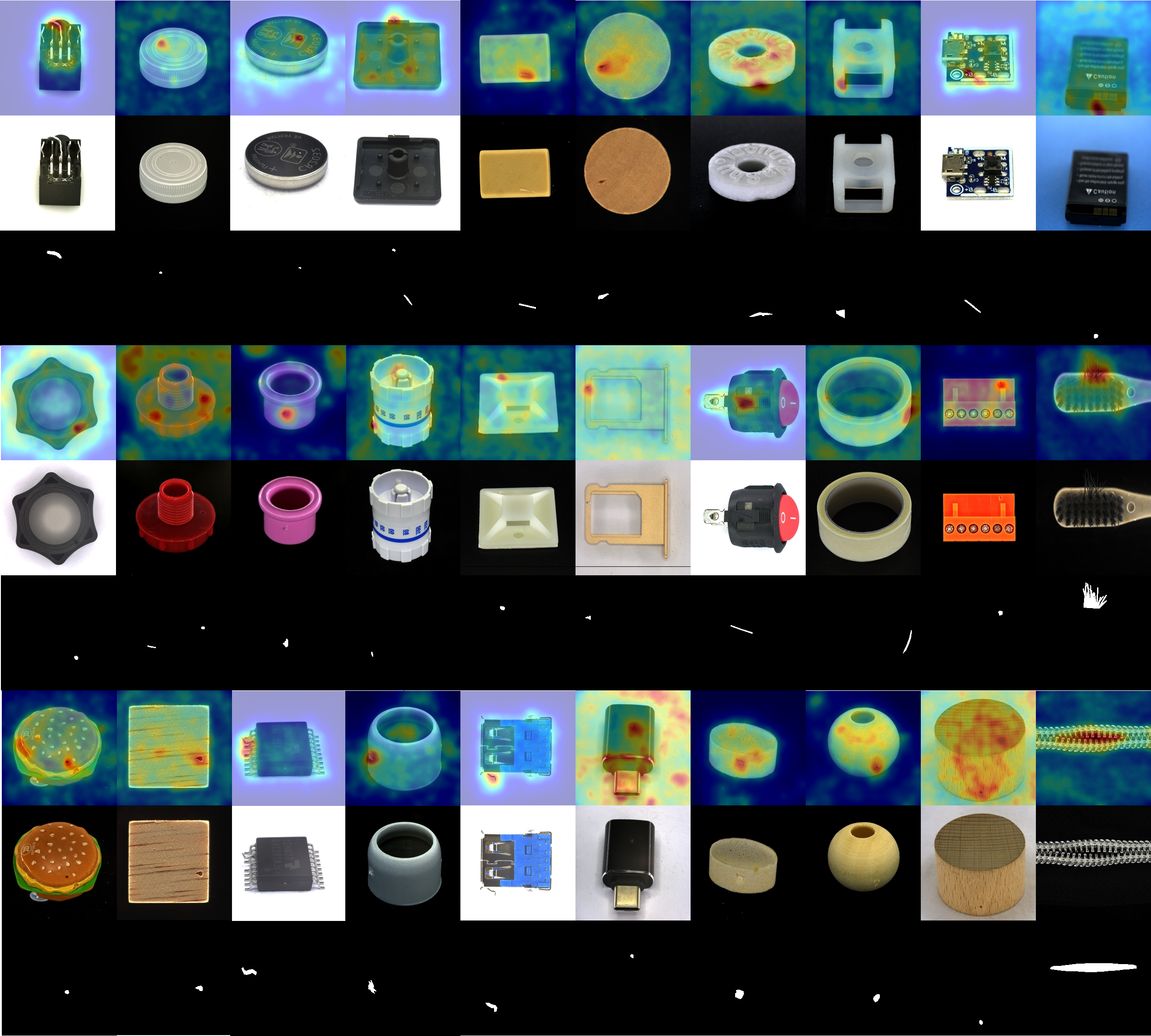}
    \caption{\textbf{Qualitative results on the Real-IAD dataset.} Top row: Predicted anomaly maps. Middle row: Original anomalous images. Bottom row: Ground-truth masks.}
    \label{fig:quali_realiad}
\end{figure}

Figure~\ref{fig:quali_mvtec} presents visualizations on the MVTec AD dataset, illustrating the effectiveness of BoRAD across diverse industrial textures and objects (e.g., pills, cables, wood, capsules). Figure~\ref{fig:quali_visa} showcases results on the VisA dataset, where our model accurately highlights anomalous features despite complex object structures and multi-object backgrounds. Finally, Figure~\ref{fig:quali_realiad} demonstrates precise pixel-level localization capabilities on the challenging high-resolution Real-IAD benchmark.

\section{Detailed Results per Category}
\label{app:detailed_results}

In this section, we provide the detailed per-category performance for MVTec AD, VisA and Real-IAD datasets. All values are reported in percentages (\%).

\newpage

\begin{sidewaystable}[p]
  \centering
  \caption{\textbf{Detailed Anomaly Detection results on MVTec AD and VisA.} Each cell contains three Image-level metrics: AU-ROC ($\uparrow$) / AP ($\uparrow$) / F1-max ($\uparrow$). All values are rounded to one decimal place. The best result for each individual metric in a row is highlighted in \textbf{bold}.}
  \label{tbl:full_detection_results}
  \vspace{5pt}
  \resizebox{\textheight}{!}{
  \begin{tabular}{l|cccccccccc}
    \toprule
    \textbf{Category} & \textbf{UniAD} & \textbf{RD4AD} & \textbf{SimpleNet} & \textbf{DeSTSeg} & \textbf{DiAD} & \textbf{HVQ-Trans} & \textbf{MambaAD} & \textbf{ViTAD} & \textbf{OmiAD} & \textbf{BoRAD (Ours)} \\
    \midrule
    \multicolumn{11}{c}{\textbf{MVTec AD Dataset}} \\
    \midrule
    Bottle & 99.7/\textbf{100.0/100.0} & 99.6/99.9/98.4 & \textbf{100.0/100.0/100.0} & 98.7/99.6/96.8 & 99.7/96.5/91.8 & \textbf{100.0/100.0/100.0} & \textbf{100.0/100.0/100.0} & \textbf{100.0/100.0/100.0} & \textbf{100.0/100.0/100.0} & \textbf{100.0/100.0/100.0} \\
    Cable & 95.2/95.9/88.0 & 84.1/89.5/82.5 & 97.5/98.5/94.7 & 89.5/94.6/85.9 & 94.8/98.8/95.2 & 99.0/98.8/95.1 & 98.8/99.2/\textbf{95.7} & \textbf{99.7/99.9}/95.7 & 98.4/\textbf{99.4}/95.6 & 97.6/98.5/94.2 \\
    Capsule & 86.9/97.8/94.4 & 94.1/96.9/\textbf{96.9} & 90.7/97.9/93.5 & 82.8/95.9/92.6 & 89.0/97.5/95.5 & 95.4/99.2/96.3 & 94.4/98.7/94.9 & \textbf{100.0/100.0}/100.0 & 94.7/99.3/96.8 & 98.9/99.8/97.7 \\
    Hazelnut & 99.8/\textbf{100.0}/99.3 & 60.8/69.8/86.4 & 99.9/99.9/99.3 & 98.8/99.2/98.6 & 99.5/99.7/97.3 & \textbf{100.0/100.0}/99.3 & \textbf{100.0/100.0/100.0} & 99.8/99.9/98.6 & \textbf{100.0/100.0/100.0} & \textbf{100.0/100.0/100.0} \\
    Metal Nut & 99.2/99.9/\textbf{99.5} & \textbf{100.0/100.0/99.5} & 96.9/99.3/96.1 & 92.9/98.4/92.2 & 99.1/96.0/91.6 & 99.9/99.9/98.9 & 99.9/\textbf{100.0}/99.5 & 99.7/99.9/98.4 & 99.4/99.9/98.9 & \textbf{100.0/100.0/100.0} \\
    Pill & 93.7/98.7/95.7 & \textbf{97.5/99.6/96.8} & 88.2/97.7/92.5 & 77.1/94.4/91.7 & 95.7/98.5/94.5 & 95.8/99.2/94.9 & 97.0/99.5/96.2 & \textbf{100.0/100.0/100.0} & 94.2/99.2/95.4 & 98.5/99.7/97.8 \\
    Screw & 87.5/96.5/89.0 & \textbf{97.7}/99.3/95.8 & 76.7/90.6/87.7 & 69.9/88.4/85.4 & 90.7/\textbf{99.7/97.9} & 95.6/97.9/92.1 & 94.7/97.9/94.0 & \textbf{98.5}/99.1/95.7 & 96.9/98.8/96.3 & 97.7/99.2/96.4 \\
    Toothbrush & 94.2/97.4/95.2 & 97.2/99.0/94.7 & 89.7/95.7/92.3 & 71.7/89.3/84.5 & \textbf{99.7}/99.9/99.2 & 93.6/99.9/98.4 & 98.3/99.3/98.4 & 98.9/99.6/96.8 & \textbf{99.7/100.0/100.0} & 97.2/98.9/95.8 \\
    Transistor & 99.8/98.0/93.8 & 94.2/95.2/90.0 & 99.2/98.7/97.6 & 78.2/79.5/68.8 & 99.8/99.6/97.4 & 99.7/99.5/96.4 & \textbf{100.0/100.0}/98.6 & 98.8/98.3/92.5 & 99.9/99.9/98.8 & 99.1/98.6/95.9 \\
    Zipper & 95.8/99.5/97.1 & 99.5/99.9/99.2 & 99.0/99.7/98.3 & 88.4/96.3/93.1 & 95.1/99.1/94.4 & 97.9/99.6/98.3 & 99.3/99.8/97.5 & 97.6/99.3/97.1 & 99.8/\textbf{100.0/99.6} & \textbf{99.9/100.0}/99.4 \\
    Carpet & 99.8/99.9/99.4 & 98.5/99.6/97.2 & 95.7/98.7/93.2 & 95.9/98.8/94.9 & 99.4/99.9/98.3 & \textbf{99.9/100.0/100.0} & 99.8/99.9/99.4 & 99.5/99.9/99.4 & 99.6/\textbf{100.0}/99.4 & 98.6/99.6/97.4 \\
    Grid & 98.2/99.5/97.3 & 98.0/99.4/96.5 & 97.6/99.2/96.4 & 97.9/99.2/96.6 & 98.5/99.8/97.7 & 97.0/99.5/97.3 & \textbf{100.0/100.0}/99.1 & 91.3/97.0/93.0 & 99.8/99.9/99.1 & 99.8/99.9/98.6 \\
    Leather & \textbf{100.0/100.0/100.0} & \textbf{100.0/100.0/100.0} & \textbf{100.0/100.0/100.0} & 99.2/99.8/98.9 & 99.8/99.7/97.6 & \textbf{100.0/100.0/100.0} & \textbf{100.0/100.0/100.0} & \textbf{100.0/100.0/100.0} & \textbf{100.0/100.0/100.0} & \textbf{100.0/100.0/100.0} \\
    Tile & 99.3/99.8/98.2 & 98.3/99.3/96.4 & 99.3/99.8/\textbf{98.8} & 97.0/98.9/95.3 & 96.8/\textbf{99.9}/98.4 & 99.2/99.8/98.2 & 98.2/99.3/95.4 & \textbf{100.0/100.0}/92.5 & \textbf{100.0}/99.9/98.8 & \textbf{100.0/100.0}/100.0 \\
    Wood & 98.6/99.6/96.6 & 99.2/99.8/98.3 & 98.4/99.5/96.7 & \textbf{99.9/100.0}/99.2 & 99.7/\textbf{100.0/100.0} & 97.2/99.6/97.4 & 98.8/99.6/96.6 & 97.6/99.3/97.1 & 99.0/99.8/98.3 & 99.3/99.8/98.3 \\
    \midrule
    \textbf{Mean} & 96.5/98.8/96.2 & 94.6/96.5/95.2 & 95.3/98.4/95.8 & 89.2/95.5/91.6 & 97.2/99.0/96.5 & 98.0/99.5/97.5 & 98.6/99.6/97.8 & 98.3/99.4/97.3 & 98.8/\textbf{99.7/98.5} & \textbf{99.1}/99.6/98.1 \\
    \midrule
    \multicolumn{11}{c}{\textbf{VisA Dataset}} \\
    \midrule
    PCB1 & 92.8/92.7/87.8 & 96.2/95.5/91.9 & 91.6/91.9/86.0 & 87.6/83.1/83.7 & 88.1/88.7/80.7 & 96.7/93.2/87.7 & 95.4/93.0/91.6 & 95.8/94.7/91.8 & \textbf{97.8/97.4/95.1} & 97.2/97.3/93.5 \\
    PCB2 & 87.8/87.7/83.1 & \textbf{97.8}/97.8/\textbf{94.2} & 92.4/93.3/84.5 & 86.5/85.8/82.6 & 91.4/91.4/84.7 & 93.4/94.8/88.0 & 94.2/93.7/89.3 & 90.6/89.9/85.3 & \textbf{97.8/\textbf{98.5}/94.1} & \textbf{97.6}/97.7/\textbf{95.1} \\
    PCB3 & 78.6/78.6/76.1 & 96.4/\textbf{96.2/91.0} & 89.1/91.1/82.6 & 93.7/95.1/87.0 & 86.2/87.6/77.6 & 92.0/87.1/79.5 & 93.7/94.1/86.7 & 90.9/91.2/83.9 & 96.7/95.1/87.6 & \textbf{97.5/97.8/92.9} \\
    PCB4 & 98.8/98.8/94.3 & 99.9/99.9/\textbf{99.0} & 97.0/97.0/93.5 & 97.8/97.8/92.7 & 99.6/99.5/97.0 & 99.5/99.0/97.0 & 99.9/99.9/98.5 & 99.1/98.9/96.6 & \textbf{100.0/100.0}/99.0 & 99.9/99.9/99.0 \\
    Macaroni 1 & 79.9/79.8/72.7 & 75.9/61.5/76.8 & 85.9/82.5/73.1 & 76.6/69.0/71.0 & 85.7/85.2/78.8 & 93.1/84.1/79.8 & 91.6/89.8/81.6 & 85.8/83.9/76.7 & \textbf{97.3/97.5/92.8} & 94.8/93.3/89.0 \\
    Macaroni 2 & 71.6/71.6/69.9 & \textbf{88.3/84.5/83.8} & 68.3/54.3/59.7 & 68.9/62.1/67.7 & 62.5/57.4/69.6 & 86.2/84.1/81.5 & 81.6/78.0/73.8 & 79.1/74.7/74.9 & 85.1/83.3/79.5 & 85.5/82.0/78.8 \\
    Capsules & 55.6/55.6/76.9 & 82.2/90.4/81.3 & 74.1/82.8/74.6 & \textbf{87.1/93.0/84.2} & 58.2/69.0/78.5 & 77.1/83.4/77.5 & \textbf{91.8/95.0/88.8} & 79.2/87.6/79.8 & 85.7/89.0/78.8 & 88.9/94.1/85.8 \\
    Candles & 94.1/94.0/86.1 & 92.3/92.9/86.0 & 84.1/73.3/76.6 & 94.9/94.8/89.2 & 92.8/92.0/87.6 & 96.8/98.0/93.1 & 96.8/96.9/90.1 & 90.4/91.2/83.7 & \textbf{97.4/98.6/93.4} & 95.2/95.4/89.9 \\
    Cashew & 92.8/92.8/91.4 & 92.0/95.8/90.7 & 88.0/91.3/84.7 & 92.0/96.1/88.1 & 91.5/95.7/89.7 & 94.9/96.8/90.4 & 94.5/97.3/91.1 & 87.8/94.2/86.1 & 93.3/96.4/90.9 & \textbf{95.9/98.1/93.2} \\
    Chewing gum & 96.3/96.2/95.2 & 94.9/97.5/92.1 & 96.4/98.2/93.8 & 95.8/98.3/94.7 & 99.1/99.5/95.9 & \textbf{99.4/99.6/97.5} & 97.7/98.9/94.2 & 94.9/97.7/91.4 & 99.2/\textbf{99.8}/97.5 & 98.2/99.2/95.5 \\
    Fryum & 83.0/83.0/85.0 & 95.3/97.9/91.5 & 88.4/93.0/83.3 & 92.1/96.1/89.5 & 89.8/95.0/87.2 & 90.4/94.5/84.9 & 95.2/97.7/90.5 & 94.3/97.4/90.9 & 94.0/96.5/88.5 & \textbf{95.6/98.0/92.0} \\
    Pipe fryum & 94.7/94.7/93.9 & 97.9/98.9/96.5 & 90.8/95.5/88.6 & 94.1/97.1/91.9 & 96.2/98.1/93.7 & 98.5/98.4/94.0 & 98.7/99.3/97.0 & 97.8/99.0/94.7 & 98.9/99.4/97.0 & \textbf{99.7/99.8/98.4} \\
    \midrule
    \textbf{Mean} & 85.5/85.5/84.4 & 92.4/92.4/89.6 & 87.2/87.0/81.8 & 88.9/89.0/85.2 & 86.8/88.3/85.1 & 93.2/92.8/87.6 & 94.3/94.5/89.4 & 90.5/91.7/86.3 & 95.3/96.0/91.2 & \textbf{95.5/96.1/91.9} \\
    \bottomrule
  \end{tabular}}
\end{sidewaystable}

\begin{sidewaystable}[p]
  \centering
  \caption{\textbf{Detailed Anomaly Localization results on MVTec AD and VisA.} Each cell contains four Pixel-level metrics: AU-ROC ($\uparrow$) / AP ($\uparrow$) / F1-max ($\uparrow$) / AU-PRO ($\uparrow$). All values are rounded to one decimal place. The best result for each individual metric in a row is highlighted in \textbf{bold}.}
  \label{tbl:full_localization_results}
  \vspace{5pt}
  \resizebox{\textheight}{!}{
  \begin{tabular}{l|cccccccccc}
    \toprule
    \textbf{Category} & \textbf{UniAD} & \textbf{RD4AD} & \textbf{SimpleNet} & \textbf{DeSTSeg} & \textbf{DiAD} & \textbf{HVQ-Trans} & \textbf{MambaAD} & \textbf{ViTAD} & \textbf{OmiAD} & \textbf{BoRAD (Ours)} \\
    \midrule
    \multicolumn{11}{c}{\textbf{MVTec AD Dataset}} \\
    \midrule
    Bottle & 98.1/66.0/69.2/93.1 & 97.8/68.2/67.6/94.0 & 97.2/53.8/62.4/89.0 & 93.3/61.7/56.0/67.5 & 98.4/52.2/54.8/86.6 & 98.3/71.8/70.2/94.6 & 98.8/\textbf{79.7/76.7}/95.2 & \textbf{99.0}/60.5/64.1/94.7 & 98.6/74.9/73.8/\textbf{95.6} & 98.6/75.2/73.8/\textbf{95.6} \\
    Cable & 97.3/39.9/45.2/86.1 & 85.1/26.3/33.6/75.1 & 96.7/42.4/51.2/85.4 & 89.3/37.5/40.5/49.4 & 96.8/50.1/57.8/80.5 & 98.1/\textbf{52.4/59.0}/87.5 & 95.8/42.2/48.1/90.3 & \textbf{98.6}/31.2/36.7/\textbf{95.8} & 98.3/60.5/62.9/91.8 & 97.0/46.6/53.2/91.3 \\
    Capsule & 98.5/42.7/46.5/92.1 & 98.8/43.4/50.0/94.8 & 98.5/5.4/44.3/84.5 & 95.8/47.9/48.9/62.1 & 97.1/42.0/45.3/87.2 & 98.8/45.3/49.7/90.7 & 98.4/43.9/47.7/92.6 & 98.3/42.7/47.8/92.0 & 98.9/48.1/52.0/92.5 & \textbf{99.0/49.6/52.5/96.0} \\
    Hazelnut & 98.1/55.2/56.8/94.1 & 97.9/36.2/51.6/92.7 & 98.4/44.6/51.4/87.4 & 98.2/65.8/61.6/84.5 & 98.3/\textbf{79.2/80.4}/91.5 & 98.8/62.7/63.2/92.5 & \textbf{99.0}/63.6/64.4/95.7 & \textbf{99.0}/64.6/64.0/95.2 & 98.6/59.6/59.9/94.3 & 98.9/59.9/61.0/\textbf{96.4} \\
    Metal Nut & 62.7/14.6/29.2/81.8 & 94.8/55.5/66.4/91.9 & \textbf{98.0/83.1}/79.4/85.2 & 84.2/42.0/22.8/53.0 & 97.3/30.0/38.3/90.6 & 96.3/67.1/75.5/90.9 & 96.7/74.5/79.1/93.7 & 96.4/75.1/77.3/92.4 & 96.5/66.6/75.6/90.3 & 97.0/74.8/\textbf{80.4/94.3} \\
    Pill & 95.0/44.0/53.9/95.3 & 97.5/63.4/65.2/95.8 & 96.5/72.4/67.7/81.9 & 96.2/61.7/41.8/27.9 & 95.7/46.0/51.4/89.0 & 97.1/50.1/57.6/94.9 & 97.4/64.0/66.5/95.7 & \textbf{98.7/77.8/75.2}/95.3 & 96.6/56.8/60.7/95.9 & 98.0/69.1/68.8/\textbf{96.4} \\
    Screw & 98.3/28.7/37.6/95.2 & 99.4/40.2/44.6/96.8 & 96.5/15.9/23.2/84.0 & 93.8/19.9/25.3/47.3 & 97.9/\textbf{60.6/59.6}/95.0 & 98.9/28.8/36.2/94.3 & \textbf{99.5}/49.8/50.9/97.1 & 99.0/34.0/41.0/93.5 & \textbf{99.5}/38.7/43.5/97.2 & 98.8/44.9/47.8/\textbf{97.4} \\
    Toothbrush & 98.4/34.9/45.7/87.9 & 99.0/53.6/58.8/92.0 & 98.4/46.9/52.5/87.4 & 96.2/52.9/58.8/30.9 & 99.0/\textbf{78.7/72.8}/95.0 & 98.6/40.8/51.4/89.2 & 99.0/48.5/59.2/91.7 & \textbf{99.1}/51.3/61.9/90.9 & 98.7/40.5/56.2/91.1 & 99.0/54.4/60.8/\textbf{91.6} \\
    Transistor & 97.9/59.5/64.6/93.5 & 85.9/42.3/45.2/74.7 & 95.8/58.2/56.0/83.2 & 73.6/38.4/39.2/43.9 & 95.1/15.6/31.7/90.0 & 97.9/71.2/67.2/95.4 & 96.5/69.4/67.1/87.0 & 93.9/58.4/55.3/76.8 & 98.4/\textbf{73.4/72.5/96.1} & 91.6/51.6/54.2/83.2 \\
    Zipper & 96.8/40.1/49.9/92.6 & \textbf{98.5}/53.9/60.3/94.1 & 97.9/53.4/54.6/90.7 & 97.3/64.7/59.2/66.9 & 96.2/60.7/60.0/91.6 & 97.5/38.7/48.8/91.7 & 98.4/60.4/61.7/94.3 & 97.6/99.3/97.1 & 98.6/52.7/59.3/\textbf{95.6} & 98.4/55.2/59.9/95.2 \\
    Carpet & 98.5/49.9/51.1/94.4 & 99.0/58.5/60.4/95.1 & 97.4/38.7/43.2/90.6 & 93.6/59.9/58.9/89.3 & 98.6/42.2/46.4/90.6 & 98.7/57.5/57.7/94.7 & \textbf{99.2}/60.0/63.3/\textbf{96.7} & 99.5/99.9/99.4 & 98.5/52.9/54.8/94.6 & 99.0/56.8/61.1/95.5 \\
    Grid & 63.1/0.7/1.9/92.9 & 96.5/23.0/28.4/97.0 & 96.8/20.5/27.6/88.6 & 97.0/42.1/46.9/86.8 & 96.6/\textbf{66.0/64.1}/94.0 & 97.0/24.5/30.5/89.5 & \textbf{99.2}/47.4/47.7/97.0 & 99.7/\textbf{100.0}/99.1 & 98.5/35.4/37.1/95.5 & \textbf{99.2}/46.5/48.8/\textbf{97.2} \\
    Leather & 98.8/32.9/34.4/96.8 & 99.3/38.0/45.1/97.4 & 98.7/28.5/32.9/92.7 & 99.5/\textbf{71.5/66.5}/91.1 & 98.8/56.1/62.3/91.3 & 98.8/33.7/36.6/97.6 & 99.4/50.3/53.3/\textbf{98.7} & \textbf{100.0/100.0/100.0} & 98.9/36.3/39.4/96.9 & 99.4/44.9/49.6/97.7 \\
    Tile & 91.8/42.1/50.6/78.4 & 95.3/48.5/60.5/85.8 & 95.7/60.5/59.9/\textbf{90.6} & 93.0/\textbf{71.0/66.2}/87.1 & 92.4/65.7/64.1/\textbf{90.7} & 92.2/41.6/52.9/81.2 & 93.8/45.1/54.8/80.0 & \textbf{100.0/100.0}/100.0 & 92.7/47.5/54.7/82.2 & \textbf{96.2}/53.8/63.9/88.2 \\
    Wood & 93.2/37.2/41.5/86.7 & 95.3/47.8/51.0/90.0 & 91.4/34.8/39.7/76.3 & 95.9/\textbf{77.3/71.3}/83.4 & 93.3/43.3/43.5/\textbf{97.5} & 92.4/37.2/42.6/86.6 & 94.4/46.2/48.2/91.2 & 98.7/99.6/96.7 & 94.2/44.5/48.0/88.5 & 95.3/51.4/51.6/90.1 \\
    \midrule
    \textbf{Mean} & 96.8/43.4/49.5/90.7 & 96.1/48.6/53.8/91.1 & 96.9/45.9/49.7/86.5 & 93.1/54.3/50.9/64.8 & 96.8/52.6/55.5/90.7 & 97.3/48.2/53.3/91.4 & 97.7/\textbf{56.3}/59.2/93.1 & 98.3/99.4/97.3 & 97.7/52.6/56.7/93.2 & \textbf{97.8}/56.0/\textbf{59.4/93.8} \\
    \midrule
    \multicolumn{11}{c}{\textbf{VisA Dataset}} \\
    \midrule
    PCB1 & 93.3/3.9/8.3/64.1 & 99.4/66.2/62.4/95.8 & 99.2/86.1/78.8/83.6 & 95.8/46.4/49.0/83.2 & 98.7/49.6/52.8/80.2 & 99.4/63.1/58.8/87.4 & \textbf{99.8}/77.1/72.4/92.8 & 99.5/64.5/61.7/89.6 & 99.7/70.8/66.2/93.0 & 99.7/\textbf{80.5/74.6/95.9} \\
    PCB2 & 93.9/4.2/9.2/66.9 & 98.0/\textbf{22.3/30.0}/90.8 & 96.6/8.9/18.6/85.7 & 97.3/14.6/28.2/79.9 & 95.2/7.5/16.7/67.0 & 98.0/10.1/18.2/82.0 & \textbf{98.9}/13.3/23.4/89.6 & 97.9/12.6/21.2/82.0 & \textbf{98.9}/16.6/22.2/87.1 & \textbf{98.8}/19.1/28.5/\textbf{92.6} \\
    PCB3 & 97.3/13.8/21.9/70.6 & 97.9/26.2/35.2/\textbf{93.9} & 97.2/\textbf{31.0/36.1}/85.1 & 97.7/28.1/33.4/62.4 & 96.7/8.0/18.8/68.9 & 98.3/21.1/23.7/80.5 & \textbf{99.1}/18.3/27.4/89.1 & 98.2/22.4/26.4/88.0 & \textbf{99.1}/29.7/30.4/87.2 & 98.9/28.3/30.5/93.8 \\
    PCB4 & 94.9/14.7/22.9/72.3 & 97.8/31.4/37.0/88.7 & 93.9/23.9/32.9/61.1 & 95.8/\textbf{53.0/53.2}/76.9 & 97.0/17.6/27.2/85.0 & 97.7/21.1/29.8/85.9 & 98.6/47.0/46.9/87.6 & \textbf{99.1}/42.9/48.3/\textbf{91.8} & 98.1/42.0/44.1/86.4 & 98.3/41.9/43.1/89.9 \\
    Macaroni 1 & 97.4/3.7/9.7/84.0 & 99.4/2.9/6.9/95.3 & 98.9/3.5/8.4/92.0 & 99.1/5.8/13.4/62.4 & 94.1/10.2/16.7/68.5 & 99.4/9.9/19.3/91.2 & 99.5/17.5/27.6/95.2 & 98.5/8.0/19.3/89.2 & \textbf{99.7/20.1/29.8}/96.3 & \textbf{99.7/22.5/31.3/96.7} \\
    Macaroni 2 & 95.2/0.9/4.3/76.6 & \textbf{99.7/13.2/21.8/97.4} & 93.2/0.6/3.9/77.8 & 98.5/6.3/14.4/70.0 & 93.6/0.9/2.8/73.1 & 98.5/5.5/13.8/91.3 & 99.5/9.2/16.1/96.2 & 98.1/3.6/10.4/87.2 & 99.4/8.5/15.4/94.0 & 99.5/10.2/18.5/96.6 \\
    Capsules & 88.7/3.0/7.4/43.7 & 99.4/60.4/60.8/93.1 & 97.1/52.9/53.3/73.7 & 96.9/33.2/9.1/76.7 & 97.3/10.0/21.0/77.9 & 99.0/51.9/55.2/76.3 & 99.1/61.3/59.8/91.8 & 98.2/30.4/41.4/75.1 & 99.4/62.7/59.6/90.8 & \textbf{99.6/68.1/64.8/94.8} \\
    Candles & 98.5/17.6/27.9/91.6 & 99.1/25.3/35.8/94.9 & 97.6/8.4/16.5/87.6 & 98.7/\textbf{39.9/45.8}/69.0 & 97.3/12.8/22.8/89.4 & 99.2/20.4/30.6/92.7 & 99.0/23.2/32.4/\textbf{95.5} & 96.2/16.8/26.4/85.2 & \textbf{99.4}/25.7/35.1/97.1 & 99.2/26.3/36.7/95.4 \\
    Cashew & 98.6/51.7/58.3/87.9 & 91.7/44.2/49.7/86.2 & \textbf{98.9/68.9/66.0}/84.1 & 87.9/47.6/52.1/66.3 & 90.9/53.1/60.9/61.8 & 99.2/58.3/60.9/89.3 & 94.3/46.8/51.4/87.8 & 98.5/63.9/62.7/78.8 & 97.2/46.3/55.3/83.8 & 95.6/57.2/58.3/\textbf{90.1} \\
    Chewing gum & \textbf{98.8}/54.9/56.1/81.3 & 98.7/59.9/61.7/76.9 & 97.9/26.8/29.8/78.3 & 98.8/86.9/81.0/68.3 & 94.7/11.9/25.8/59.5 & 98.8/43.7/44.4/81.7 & 98.1/57.5/59.9/79.7 & 97.8/61.6/58.7/71.5 & \textbf{98.8}/58.0/55.6/73.1 & 98.7/61.5/63.8/\textbf{82.1} \\
    Fryum & 95.9/34.0/40.6/76.2 & 97.0/47.6/51.5/\textbf{93.4} & 93.0/39.1/45.4/85.1 & 88.1/35.2/38.5/47.7 & \textbf{97.6/58.6/60.1}/81.3 & 97.7/50.8/55.3/83.6 & 96.9/47.8/51.9/91.6 & 97.5/47.1/50.3/87.8 & 97.7/47.7/55.2/87.2 & 97.0/47.8/52.0/92.2 \\
    Pipe fryum & 98.9/50.2/57.7/91.5 & 99.1/56.8/58.8/95.4 & 98.5/65.6/63.4/83.0 & 98.9/\textbf{78.8/72.7}/45.9 & 99.4/72.7/69.9/89.9 & 99.4/64.6/65.8/93.4 & 99.1/53.5/58.5/95.1 & \textbf{99.5}/66.0/66.5/94.7 & 99.2/56.2/60.1/94.5 & 99.1/56.6/58.2/\textbf{95.8} \\
    \midrule
    \textbf{Mean} & 95.9/21.0/27.0/75.6 & 98.1/38.0/42.6/91.8 & 96.8/34.7/37.8/81.4 & 96.1/39.6/43.4/67.4 & 96.0/26.1/33.0/75.2 & 98.7/35.0/39.6/86.3 & 98.5/39.4/44.0/91.0 & 98.2/36.6/41.1/85.1 & \textbf{98.9}/40.4/44.1/89.2 & 98.7/\textbf{43.4/46.7/93.0} \\
    \bottomrule
  \end{tabular}}
\end{sidewaystable}

\begin{sidewaystable}[p]
  \centering
  \caption{\textbf{Detailed Anomaly Detection results on the Real-IAD dataset.} Each cell contains three Image-level metrics: AU-ROC ($\uparrow$) / AP ($\uparrow$) / F1-max ($\uparrow$). Values are rounded to one decimal place. The best result for each metric is highlighted in \textbf{bold}.}
  \label{tbl:realiad_detection_detail}
  \vspace{5pt}
  \setlength{\tabcolsep}{3pt}
  \resizebox{\textheight}{!}{
  \begin{tabular}{l|cccccccccc}
    \toprule
    \textbf{Category} & \textbf{UniAD} & \textbf{RD4AD} & \textbf{SimpleNet} & \textbf{DeSTSeg} & \textbf{DiAD} & \textbf{HVQ-Trans} & \textbf{MambaAD} & \textbf{ViTAD} & \textbf{OmiAD} & \textbf{BoRAD (Ours)} \\
    \midrule
    audiojack & 81.4/76.6/64.9 & 76.2/63.2/60.8 & 58.4/44.2/50.9 & 81.1/72.6/64.5 & 76.5/54.3/65.7 & 81.0/80.1/74.5 & 84.2/76.5/67.4 & -- & \textbf{84.5/82.0/77.8} & 81.9/75.1/63.9 \\
    bottlecap & 92.5/91.7/81.7 & 89.5/86.3/81.0 & 54.1/47.6/60.3 & 78.1/74.6/68.1 & 91.6/\textbf{94.0/87.9} & 89.0/87.6/77.3 & 92.8/92.0/82.1 & -- & \textbf{93.7}/92.7/85.2 & \textbf{93.7}/92.6/83.5 \\
    buttonbattery & 75.9/81.6/76.3 & 73.3/78.9/76.1 & 52.5/60.5/72.4 & \textbf{86.7/89.2/83.5} & 80.5/71.3/70.6 & 82.2/88.5/78.9 & 79.8/85.3/77.8 & -- & 84.9/\textbf{90.1}/80.1 & 83.6/87.4/80.3 \\
    endcap & 80.9/86.1/78.0 & 79.8/84.0/77.8 & 51.6/60.8/72.9 & 77.9/81.1/77.1 & \textbf{85.1}/83.4/\textbf{84.8} & 79.7/85.2/79.4 & 78.0/82.8/77.2 & -- & 79.4/80.4/80.8 & 83.0/\textbf{86.8}/79.6 \\
    eraser & 90.3/89.2/80.2 & 90.0/88.7/79.7 & 46.4/39.1/55.8 & 84.6/82.9/71.8 & 80.0/80.0/77.3 & 89.2/89.7/81.9 & 87.5/86.2/76.1 & -- & 89.5/\textbf{90.2/84.2} & \textbf{91.0}/89.6/80.6 \\
    firehood & 80.6/74.8/66.4 & 78.3/70.1/64.5 & 58.1/41.9/54.4 & 81.7/72.4/67.7 & 83.3/81.7/80.5 & 93.1/85.4/83.1 & 79.3/72.5/64.8 & -- & \textbf{94.1/87.6/83.3} & 80.5/74.4/66.8 \\
    mint & 67.0/66.6/64.6 & 65.8/63.1/64.8 & 52.4/50.3/63.7 & 58.4/55.8/63.7 & \textbf{76.7/76.7/76.0} & 63.0/75.7/75.1 & 70.1/70.8/65.5 & -- & 66.0/\textbf{77.7/75.0} & 72.5/73.4/66.0 \\
    mounts & 87.6/77.3/77.2 & 88.6/79.9/74.8 & 58.7/48.1/52.4 & 74.7/56.5/63.1 & 75.3/74.5/\textbf{82.5} & 92.7/88.2/81.2 & 86.8/78.0/73.5 & -- & \textbf{95.2/92.3/85.9} & 89.5/81.2/76.8 \\
    pcb & 81.0/88.2/79.1 & 79.5/85.8/79.7 & 54.5/66.0/75.5 & 82.0/88.7/79.6 & 86.0/85.1/85.4 & 86.6/92.4/82.7 & 89.1/93.7/84.0 & -- & \textbf{92.2/95.7/87.3} & 89.5/93.6/84.4 \\
    phonebattery & 83.6/80.0/71.6 & 87.5/83.3/77.1 & 51.6/43.8/58.0 & 83.3/81.8/72.1 & 82.3/77.7/75.9 & 88.0/89.5/80.5 & 90.2/88.9/80.5 & -- & \textbf{92.6/93.0/84.5} & 91.1/88.1/80.6 \\
    plasticnut & 80.0/69.2/63.7 & 80.3/68.0/64.4 & 59.2/40.3/51.8 & 83.1/75.4/66.5 & 71.9/58.2/65.6 & 76.2/57.3/53.7 & \textbf{87.1/80.7/70.7} & -- & 84.2/67.5/62.2 & 86.7/78.7/70.5 \\
    plasticplug & 81.4/75.9/67.6 & 81.9/74.3/68.8 & 48.2/38.4/54.6 & 71.7/63.1/60.0 & 88.7/89.2/\textbf{90.9} & 92.2/92.1/84.5 & 85.7/82.2/72.6 & -- & \textbf{94.1/93.2}/86.6 & 88.7/83.5/76.0 \\
    porcelaindoll & 85.1/75.2/69.3 & \textbf{86.3}/76.3/71.5 & 66.3/54.5/52.1 & 78.7/66.2/64.3 & 72.6/66.8/65.2 & 84.7/81.5/72.9 & 88.0/82.2/74.1 & -- & 86.1/\textbf{84.5/76.3} & 85.4/77.1/69.6 \\
    regulator & 56.9/41.5/44.5 & 66.9/48.8/47.7 & 50.5/29.0/43.9 & 79.2/63.5/56.9 & 72.1/71.4/\textbf{78.2} & 69.7/27.3/37.8 & 69.7/58.7/50.4 & -- & \textbf{89.5}/69.3/67.2 & 78.8/\textbf{70.7}/60.9 \\
    rolledstripbase & 98.7/99.3/96.5 & 97.5/98.7/94.7 & 59.0/75.7/79.8 & 96.5/98.2/93.0 & 68.4/55.9/56.8 & 99.3/99.7/98.4 & 98.0/99.0/95.0 & -- & \textbf{99.8/99.9/98.9} & 99.1/99.5/97.0 \\
    simcardset & 89.7/90.3/83.2 & 91.6/91.8/84.8 & 63.1/69.7/70.8 & 95.5/96.2/89.2 & 72.6/53.7/61.5 & \textbf{97.2/98.1/92.8} & 94.4/95.1/87.2 & -- & 95.9/97.3/91.6 & 93.8/94.3/87.1 \\
    switch & 85.5/88.6/78.4 & 84.3/87.2/77.9 & 62.2/66.8/68.6 & 90.1/92.8/83.1 & 73.4/49.4/61.2 & 87.5/93.1/85.0 & 91.7/94.0/85.4 & -- & \textbf{94.8/96.9/91.5} & 93.6/94.9/87.5 \\
    tape & 97.2/96.2/89.4 & 96.0/95.1/87.6 & 49.9/41.1/54.5 & 94.5/93.4/85.9 & 73.9/57.8/66.1 & \textbf{97.6/97.3/92.8} & 96.8/95.9/89.3 & -- & \textbf{98.0/98.0/93.7} & 97.1/96.3/89.9 \\
    terminalblock & 87.5/89.1/81.0 & 89.4/89.7/83.1 & 59.8/64.7/68.8 & 83.1/86.2/76.6 & 62.1/36.4/47.8 & 95.0/96.3/90.5 & 96.1/96.8/90.0 & -- & \textbf{98.4/99.0/96.1} & 95.8/96.3/90.0 \\
    toothbrush & 78.4/80.1/75.6 & 82.0/83.8/77.2 & 65.9/70.0/70.1 & 83.7/85.3/79.0 & \textbf{91.2}/93.7/\textbf{90.9} & 87.0/92.6/84.4 & 85.1/86.2/80.3 & -- & 90.7/\textbf{94.8}/86.9 & 86.5/88.0/80.9 \\
    toy & 68.4/75.1/74.8 & 69.4/74.2/75.9 & 57.8/64.4/73.4 & 70.3/74.8/75.4 & 66.2/57.3/59.8 & 74.6/82.1/83.1 & 83.0/87.5/79.6 & -- & \textbf{89.5/93.2/87.6} & 83.7/88.0/80.2 \\
    toybrick & 77.0/71.1/66.2 & 63.6/56.1/59.0 & 58.3/49.7/58.2 & 73.2/68.7/63.3 & 68.4/45.3/55.9 & 82.5/83.7/72.6 & 70.5/63.7/61.6 & -- & \textbf{85.0/85.5/74.8} & 69.0/62.1/61.4 \\
    transistor1 & 93.7/95.9/88.9 & 91.0/94.0/85.1 & 62.2/69.2/72.1 & 90.2/92.1/84.6 & 73.1/63.1/62.7 & 93.8/97.5/91.8 & 94.4/96.0/89.0 & -- & 96.2/\textbf{98.3/92.9} & \textbf{96.3}/97.5/92.2 \\
    ublock & 88.8/84.2/75.5 & 89.5/85.0/74.2 & 62.4/48.4/51.8 & 80.1/73.9/64.3 & 75.2/68.4/67.9 & 88.5/81.1/72.9 & 89.7/85.7/75.3 & -- & 90.1/83.4/74.5 & \textbf{91.9/88.2/78.7} \\
    usb & 78.7/79.4/69.1 & 84.9/84.3/75.1 & 57.0/55.3/62.9 & 87.8/88.0/78.3 & 58.9/37.4/45.7 & 92.0/91.2/85.3 & 92.0/92.2/84.5 & -- & \textbf{95.5/94.1/90.2} & 91.8/91.3/84.0 \\
    usbadaptor & 76.8/71.3/64.9 & 71.1/61.4/62.2 & 47.5/38.4/56.5 & 80.1/74.9/67.4 & 76.9/60.2/67.2 & 77.9/75.0/69.3 & 79.4/76.0/66.3 & -- & \textbf{82.6/82.4/72.6} & 76.2/70.1/63.7 \\
    vcpill & 87.1/84.0/74.7 & 85.1/80.3/72.4 & 59.0/48.7/56.4 & 83.8/81.5/69.9 & 64.1/40.4/56.2 & 89.8/88.8/81.6 & 88.3/87.7/77.4 & -- & \textbf{91.4/90.7/82.8} & 88.7/86.9/76.4 \\
    woodenbeads & 78.4/77.2/67.8 & 81.2/78.9/70.9 & 55.1/52.0/60.2 & 82.4/78.5/73.0 & 62.1/56.4/65.9 & \textbf{79.7/84.9/76.6} & 82.5/81.7/71.8 & -- & 77.0/83.5/74.9 & \textbf{84.6}/83.3/73.9 \\
    woodstick & 80.8/72.6/63.6 & 76.9/61.2/58.1 & 58.2/35.6/45.2 & 80.4/69.2/60.3 & 74.1/66.0/62.1 & 88.9/65.4/63.2 & 80.4/69.0/63.4 & -- & \textbf{92.3}/65.9/60.5 & 79.6/65.8/60.5 \\
    zipper & 98.2/98.9/95.3 & 95.3/97.2/91.2 & 77.2/86.7/77.6 & 96.9/98.1/93.5 & 86.0/87.0/84.0 & 98.8/99.7/97.3 & 99.2/99.6/96.9 & -- & \textbf{99.8/99.9/99.0} & 98.8/99.3/95.8 \\
    \midrule
    \textbf{Mean} & 83.0/80.9/74.3 & 82.4/79.0/73.9 & 57.2/53.4/61.5 & 82.3/79.2/73.2 & 75.6/66.4/69.9 & 86.6/84.9/79.4 & 86.3/84.6/77.0 & -- & \textbf{90.1/88.6/82.8} & 87.4/85.2/78.0 \\
    \bottomrule
  \end{tabular}}
\end{sidewaystable}

\begin{sidewaystable}[p]
  \centering
  \caption{\textbf{Detailed Anomaly Localization results on the Real-IAD dataset.} Each cell contains four Pixel-level metrics: AU-ROC ($\uparrow$) / AP ($\uparrow$) / F1-max ($\uparrow$) / AU-PRO ($\uparrow$). All values are rounded to one decimal place. The best result for each individual metric in a row is highlighted in \textbf{bold}.}
  \label{tbl:realiad_localization_detail}
  \vspace{5pt}
  \setlength{\tabcolsep}{3pt}
  \resizebox{\textheight}{!}{
  \begin{tabular}{l|cccccccccc}
    \toprule
    \textbf{Category} & \textbf{UniAD} & \textbf{RD4AD} & \textbf{SimpleNet} & \textbf{DeSTSeg} & \textbf{DiAD} & \textbf{HVQ-Trans} & \textbf{MambaAD} & \textbf{ViTAD} & \textbf{OmiAD} & \textbf{BoRAD (Ours)} \\
    \midrule
    audiojack & 97.6/20.0/31.0/83.7 & 96.6/12.8/22.1/79.6 & 74.4/0.9/4.8/38.0 & 95.5/25.4/31.9/52.6 & 91.6/1.0/3.9/63.3 & 98.5/31.9/41.0/88.0 & 97.7/21.6/29.5/83.9 & -- & \textbf{99.0/47.1/51.4/91.5} & 97.5/30.2/40.5/85.1 \\
    bottlecap & 99.5/19.4/29.6/96.0 & 99.5/18.9/29.9/95.7 & 85.3/2.3/5.7/45.1 & 94.5/25.3/31.1/25.3 & 94.6/4.9/11.4/73.0 & 98.4/15.6/22.8/90.0 & \textbf{99.7/30.6/34.6/97.2} & -- & 99.4/23.4/29.0/95.2 & 99.4/25.0/33.1/96.8 \\
    buttonbattery & 96.7/28.5/34.4/77.5 & 97.6/33.8/37.8/86.5 & 75.9/3.2/6.6/40.5 & 98.3/\textbf{63.9/60.4}/36.9 & 84.1/1.4/5.3/66.9 & 99.0/58.1/59.2/85.3 & 98.1/46.7/49.5/86.2 & -- & \textbf{99.2/61.0/60.4/91.9} & 98.4/42.7/48.3/89.9 \\
    endcap & 95.8/8.8/17.4/85.4 & 96.7/12.5/22.5/89.2 & 63.1/0.5/20.8/25.7 & 89.6/14.4/22.7/29.5 & 81.3/2.0/6.9/38.2 & 95.6/6.0/14.5/84.7 & 97.0/12.0/19.6/89.4 & -- & \textbf{97.2}/9.2/14.4/\textbf{91.3} & 97.0/\textbf{15.3/23.9}/90.6 \\
    eraser & 99.3/24.4/30.9/94.1 & 99.5/30.8/36.7/96.0 & 80.6/2.7/7.1/42.8 & 95.8/\textbf{52.7/53.9}/46.7 & 91.1/7.7/15.4/67.5 & 99.2/31.6/38.4/92.4 & 99.2/30.2/38.3/93.7 & -- & \textbf{99.3}/39.5/44.2/93.3 & \textbf{99.3}/31.0/36.6/\textbf{94.9} \\
    firehood & 98.6/23.4/32.2/85.3 & 98.9/27.7/35.2/87.9 & 70.5/0.3/2.2/25.3 & 97.3/27.1/35.3/34.7 & 91.8/3.2/9.2/66.7 & \textbf{98.9/35.0/42.7}/93.3 & 98.7/25.1/31.3/86.3 & -- & \textbf{98.9/43.3/48.6/94.6} & 98.5/26.9/34.9/88.9 \\
    mint & 94.4/7.7/18.1/62.3 & 95.0/11.7/23.0/72.3 & 79.9/0.9/3.6/43.3 & 84.1/10.3/22.4/9.9 & 91.1/5.7/11.6/64.2 & 94.8/17.4/26.8/58.1 & 96.5/15.9/27.0/72.6 & -- & \textbf{96.6/29.6/38.7}/67.4 & 96.2/18.2/28.6/\textbf{77.8} \\
    mounts & 99.4/28.0/32.8/95.2 & 99.3/30.6/37.1/94.9 & 80.5/2.2/6.8/46.1 & 94.2/30.0/41.3/43.3 & 84.3/0.4/1.1/48.8 & 99.6/30.2/37.0/97.7 & 99.2/31.4/35.4/93.5 & -- & \textbf{99.7/39.2/40.0/98.6} & 99.3/32.3/37.1/94.9 \\
    pcb & 97.0/18.5/28.1/81.6 & 97.5/15.8/24.3/88.3 & 78.0/1.4/4.3/41.3 & 97.2/37.1/40.4/48.8 & 92.0/3.7/7.4/66.5 & 97.7/28.7/37.2/84.1 & \textbf{99.2}/46.3/50.4/93.1 & -- & 99.0/\textbf{48.8/51.2}/92.1 & 99.0/43.1/48.0/\textbf{93.3} \\
    phonebattery & 85.5/11.2/21.6/88.5 & 77.3/22.6/31.7/94.5 & 43.4/0.1/0.9/11.8 & 79.5/25.6/33.8/39.5 & 96.8/5.3/11.4/85.4 & 98.0/24.1/31.5/87.3 & \textbf{99.4}/36.3/41.3/95.3 & -- & 99.2/\textbf{41.1/44.5}/94.0 & 95.3/37.2/41.8/\textbf{95.4} \\
    plasticnut & 98.4/20.6/27.1/88.9 & 98.8/21.1/29.6/91.0 & 77.4/0.6/3.6/41.5 & 96.5/\textbf{44.8/45.7}/38.4 & 81.1/0.4/3.4/38.6 & 97.0/16.2/26.8/84.9 & \textbf{99.4}/33.1/37.3/\textbf{96.1} & -- & 98.3/27.2/31.1/90.1 & 99.3/28.5/33.0/95.4 \\
    plasticplug & 98.6/17.4/26.1/90.3 & 99.1/20.5/28.4/94.9 & 78.6/0.7/1.9/38.8 & 91.9/20.1/27.3/21.0 & 92.9/8.7/15.0/66.1 & 99.2/23.6/29.7/95.0 & 99.0/24.2/31.7/91.5 & -- & \textbf{99.5/37.1/41.3/97.1} & 98.9/26.6/33.5/94.3 \\
    porcelaindoll & 98.7/14.1/24.5/93.2 & 99.2/24.8/34.6/\textbf{95.7} & 81.8/2.0/6.4/47.0 & 93.1/\textbf{35.9/40.3}/24.8 & 93.1/1.4/4.8/70.4 & 97.9/11.4/18.6/89.9 & \textbf{99.2}/31.3/36.6/95.4 & -- & 98.8/18.3/26.3/93.8 & 97.9/23.5/34.1/90.6 \\
    regulator & 95.5/9.1/17.4/76.1 & 98.0/7.8/16.1/88.6 & 76.6/0.1/0.6/38.1 & 88.8/18.9/23.6/17.5 & 84.2/0.4/1.5/\textbf{78.2} & 98.0/7.0/16.2/89.7 & 97.6/20.6/29.8/87.0 & -- & \textbf{99.7/37.4/42.2/98.6} & 98.7/23.7/33.2/93.0 \\
    rolledstripbase & 99.6/20.7/32.2/97.8 & \textbf{99.7}/31.4/39.9/98.4 & 80.5/1.7/5.1/52.1 & 99.2/\textbf{48.7/50.1}/55.5 & 87.7/0.6/3.2/63.4 & 98.9/16.1/25.9/96.2 & \textbf{99.7}/37.4/42.5/98.4 & -- & \textbf{99.7}/32.4/42.5/\textbf{98.9} & \textbf{99.7}/34.2/43.2/98.8 \\
    simcardset & 97.9/31.6/39.8/85.0 & 98.5/40.2/44.2/89.5 & 71.0/6.8/14.3/30.8 & 99.1/\textbf{65.5/62.1}/73.9 & 89.9/1.7/5.8/60.4 & 99.1/39.7/43.2/93.9 & 98.8/51.1/50.6/89.4 & -- & \textbf{99.3}/48.9/50.1/\textbf{95.4} & 97.7/42.6/45.1/86.6 \\
    switch & 98.1/33.8/40.6/90.7 & 94.4/18.9/26.6/90.9 & 71.7/3.7/9.3/44.2 & 97.4/57.6/55.6/44.7 & 90.5/1.4/5.3/64.2 & 99.0/51.5/55.2/91.5 & 98.2/39.9/45.4/92.9 & -- & \textbf{99.5/63.6/63.4/95.8} & 98.0/39.4/47.0/94.2 \\
    tape & 99.7/29.2/36.9/97.5 & \textbf{99.7}/42.4/47.8/98.4 & 77.5/1.2/3.9/41.4 & 99.0/\textbf{61.7/57.6}/48.2 & 81.7/0.4/2.7/47.3 & 99.6/20.5/29.8/98.3 & \textbf{99.8}/47.1/48.2/98.0 & -- & 99.7/29.8/36.4/\textbf{98.8} & \textbf{99.7}/42.2/46.9/98.5 \\
    terminalblock & 99.2/23.1/30.5/94.4 & 99.5/27.4/35.8/97.6 & 87.0/0.8/3.6/54.8 & 96.6/40.6/44.1/34.8 & 75.5/0.1/1.1/38.5 & 99.6/35.5/39.3/97.1 & \textbf{99.8}/35.3/39.7/98.2 & -- & \textbf{99.8/48.3/51.0/98.9} & 99.7/34.9/38.7/98.4 \\
    toothbrush & 95.7/16.4/25.3/84.3 & 96.9/26.1/34.2/88.7 & 84.7/7.2/14.8/52.6 & 94.3/30.0/37.3/42.8 & 82.0/1.9/6.6/54.5 & 98.4/37.2/44.4/90.6 & 97.5/27.8/36.7/91.4 & -- & \textbf{98.8/39.8/47.9/93.8} & 97.4/34.3/40.5/91.7 \\
    toy & 93.4/4.6/12.4/70.5 & 95.2/5.1/12.8/82.3 & 67.7/0.1/0.4/25.0 & 86.3/8.1/15.9/16.4 & 82.1/1.1/4.2/50.3 & 94.2/5.4/10.4/82.2 & 96.0/16.4/25.8/86.3 & -- & \textbf{97.8/19.8/25.5/90.8} & 96.8/17.0/25.4/90.4 \\
    toybrick & 97.4/17.1/27.6/81.3 & 96.4/16.0/24.6/75.3 & 86.5/5.2/11.1/56.3 & 94.7/24.6/30.8/45.5 & 93.5/3.1/8.1/66.4 & 97.5/28.9/37.3/82.7 & 96.6/18.0/25.8/74.7 & -- & \textbf{98.6/44.3/48.7/89.6} & 96.6/21.0/27.9/78.2 \\
    transistor1 & 98.9/25.6/33.2/94.3 & 99.1/29.6/35.5/95.1 & 71.7/5.1/11.3/35.3 & 97.3/\textbf{43.8/44.5}/45.4 & 88.6/7.2/15.3/58.1 & 98.1/27.1/31.8/91.4 & \textbf{99.4}/39.4/40.0/96.5 & -- & 99.1/40.2/43.6/95.9 & \textbf{99.4}/38.3/41.2/\textbf{96.8} \\
    ublock & 99.3/22.3/29.6/94.3 & \textbf{99.6}/40.5/45.2/96.9 & 76.2/4.8/12.2/34.0 & 96.9/\textbf{57.1/55.7}/38.5 & 88.8/1.6/5.4/54.2 & 99.2/19.0/27.1/94.1 & 99.5/37.8/46.1/95.4 & -- & 99.5/24.2/35.6/\textbf{97.8} & 99.4/34.4/42.7/96.1 \\
    usb & 97.9/20.6/31.7/85.3 & 98.1/26.4/35.2/91.0 & 81.1/1.5/4.9/52.4 & 98.4/42.2/47.7/57.1 & 78.0/1.0/3.1/28.0 & 99.2/29.0/38.1/93.6 & 99.2/39.1/44.4/95.2 & -- & \textbf{99.6/43.4/48.3/97.0} & 99.0/36.7/42.6/95.1 \\
    usbadaptor & 96.6/10.5/19.0/78.4 & 94.5/9.8/17.9/73.1 & 67.9/0.2/1.3/28.9 & 94.9/\textbf{25.5/34.9}/36.4 & 94.0/2.3/06.6/75.5 & 94.5/11.8/21.1/73.0 & \textbf{97.3}/15.3/22.6/82.5 & -- & 96.8/18.1/27.3/\textbf{84.2} & 95.2/14.9/24.6/75.1 \\
    vcpill & 99.1/40.7/43.0/91.3 & 98.3/43.1/48.6/88.7 & 68.2/1.1/3.3/22.0 & 97.1/\textbf{64.7/62.3}/42.3 & 90.2/1.3/5.2/60.8 & 99.1/61.9/63.3/92.0 & 98.7/50.2/54.5/89.3 & -- & 99.0/58.4/61.2/\textbf{92.6} & 98.5/49.6/54.6/90.8 \\
    woodenbeads & 97.6/16.5/23.6/84.6 & 98.0/27.1/34.7/\textbf{85.7} & 68.1/2.4/6.0/28.3 & 94.7/\textbf{38.9/42.9}/39.4 & 85.0/1.1/4.7/45.6 & 96.6/21.5/30.0/77.2 & \textbf{98.0}/32.6/39.8/84.5 & -- & 97.3/26.2/31.4/83.1 & 97.7/32.6/39.2/85.6 \\
    woodstick & 94.0/36.2/44.3/77.2 & 97.8/30.7/38.4/85.0 & 76.1/1.4/6.0/32.0 & 97.9/\textbf{60.3/60.0}/51.0 & 90.9/2.6/8.0/60.7 & 97.8/47.3/50.3/91.4 & 97.7/40.1/44.9/82.7 & -- & \textbf{98.4}/48.5/51.9/\textbf{93.1} & 97.6/35.0/41.3/85.4 \\
    zipper & 98.4/32.5/36.1/95.1 & 99.1/44.7/50.2/96.3 & 89.9/23.3/31.2/55.5 & 98.2/35.3/39.0/78.5 & 90.2/12.5/18.8/53.5 & 98.5/37.3/43.6/94.6 & \textbf{99.3}/58.2/61.3/\textbf{97.6} & -- & 99.0/43.8/49.7/97.4 & \textbf{99.3/59.6/60.7}/97.4 \\
    \midrule
    \textbf{Mean} & 97.3/21.1/29.2/86.7 & 97.3/25.0/32.7/89.6 & 75.7/2.8/6.5/39.0 & 94.6/37.9/41.7/40.6 & 88.0/2.9/7.1/58.1 & 98.0/27.6/34.4/88.7 & 98.5/33.0/38.7/90.5 & -- & \textbf{98.9/37.7/42.6/93.1} & 98.2/32.4/38.9/\textbf{91.3} \\
    \bottomrule
  \end{tabular}}
\end{sidewaystable}
\end{document}